\documentclass[runningheads]{llncs}

\usepackage[final]{eccv}

\usepackage{eccvabbrv}

\usepackage{graphicx}
\usepackage{booktabs}
\usepackage{multirow}

\usepackage[accsupp]{axessibility}  

\usepackage{hyperref}

\usepackage{orcidlink}

\begin{document}

\title{GaussianGPT: Towards Autoregressive 3D Gaussian Scene Generation} 

\author{Nicolas von Lützow  \and
Barbara Rössle  \and 
Katharina Schmid  \and
Matthias Nießner 
}

\authorrunning{N. von Lützow et al.}

\institute{Technical University of Munich, Germany}

\maketitle
\vspace{-0.4cm}
\begin{figure}[h]
  \centering
  \includegraphics[width=\textwidth]{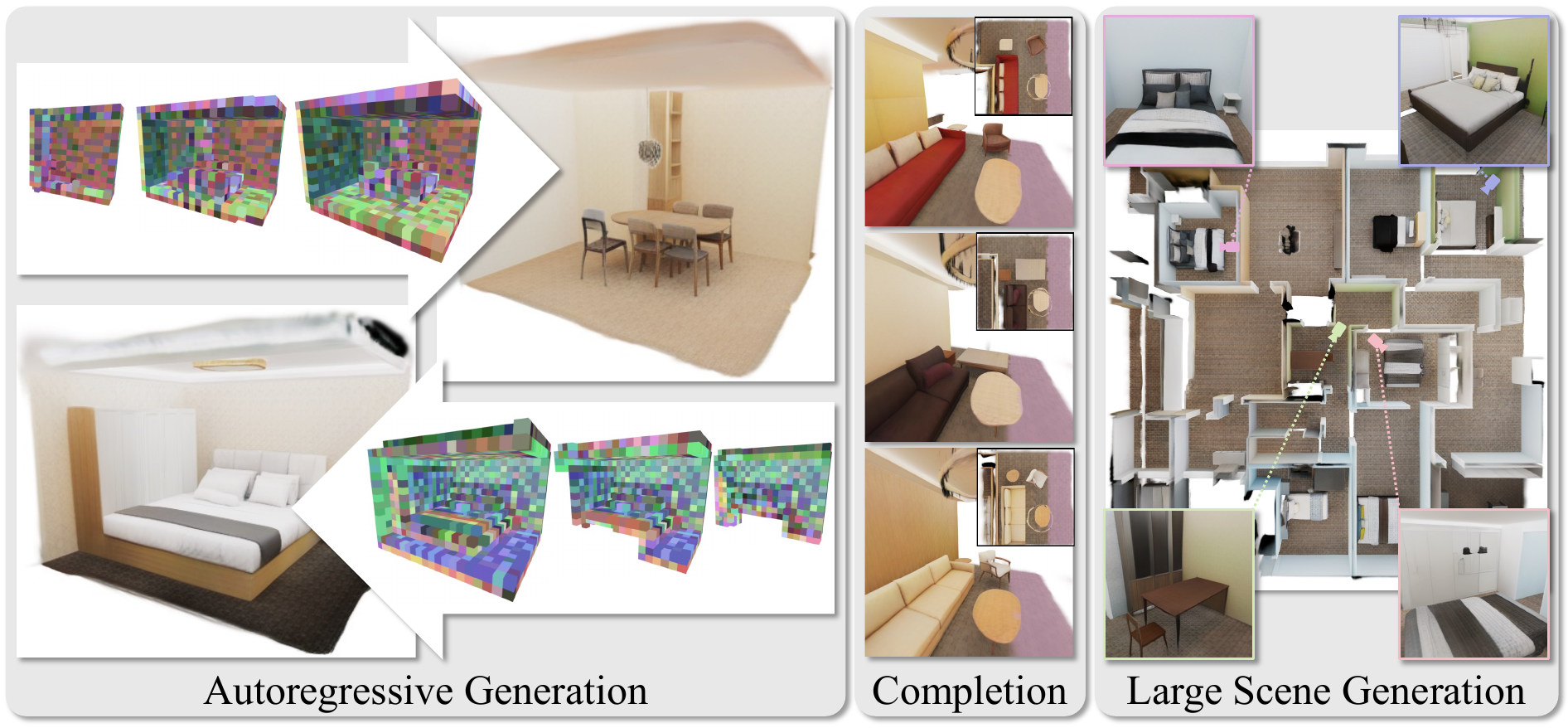} 
    \caption{
    GaussianGPT is a purely autoregressive approach for 3D Gaussian scene generation.
    Our approach enables unconditional scene generation, scene completion, and large-scale scene synthesis using only a single model.
    }
  \label{fig:teaser}
\end{figure}

\vspace{-0.8cm}
\begin{abstract}
Most recent advances in 3D generative modeling rely on diffusion or flow-matching formulations.
We instead explore a fully autoregressive alternative and introduce \textbf{GaussianGPT}, a transformer-based model that directly generates 3D Gaussians via next-token prediction, thus facilitating full 3D scene generation.
We first compress Gaussian primitives into a discrete latent grid using a sparse 3D convolutional autoencoder with vector quantization.
The resulting tokens are serialized and modeled using a causal transformer with 3D rotary positional embedding, enabling sequential generation of spatial structure and appearance.
Unlike diffusion-based methods that refine scenes holistically, our formulation constructs scenes step-by-step, naturally supporting completion, outpainting, controllable sampling via temperature, and flexible generation horizons.
This formulation leverages the compositional inductive biases and scalability of autoregressive modeling while operating on explicit representations compatible with modern neural rendering pipelines, positioning autoregressive transformers as a complementary paradigm for controllable and context-aware 3D generation.
\end{abstract}

\section{Introduction}
Recent advances in deep neural networks have enabled high-quality synthesis across images, video, audio, and language.
Extending these capabilities to structured 3D scene generation is a key step toward immersive virtual environments, embodied AI, simulation, and content creation. 
While current 3D generative approaches often rely on holistic denoising, they lack the flexibility required for the way environments are actually constructed: incrementally. In practice, 3D scenes are rarely static; they are progressively extended, completed, and edited. This creates a critical need for models that treat 3D space as a structured sequence, enabling step-by-step construction. 
Accordingly, we frame 3D generation as an autoregressive process in which the scene is iteratively built by predicting new elements conditioned on the existing spatial context.

Despite rapid progress, generating structured 3D scenes remains difficult due to the inherent high dimensionality and lack of a natural sequence in 3D data. 
Unlike 2D images, 3D environments require representations that jointly capture geometry, appearance, and semantics while maintaining multi-view consistency. 
Indoor scenes exhibit both strong regularities and high variability, demanding a balance between global structural coherence (e.g., room layout) and local compositional flexibility, including varied object placement, geometry, and appearance. 
Furthermore, the lack of a canonical ordering makes it difficult to apply successful autoregressive paradigms from other domains, as models need to serialize the 3D scene into a sequence while preserving spatial dependencies.

Recent state-of-the-art approaches \cite{xiang_native_2025, xiang_structured_2025, lan_gaussiananything_2025, roessle_l3dg_2024} predominantly rely on diffusion or flow-based models operating over neural fields, Gaussian primitives, or latent scene representations. While these methods achieve impressive visual fidelity, they typically formulate generation as a global denoising or refinement process, which can make incremental editing and structured completion less natural. In contrast, autoregressive transformers have demonstrated strong capabilities for structured sequence modeling and controllable generation across language and vision domains, yet remain comparatively underexplored for structured 3D scene synthesis. 

In this work, we introduce \textbf{GaussianGPT}, a novel method for autoregressive indoor scene generation and completion that formulates 3D synthesis as sequential prediction over structured scene primitives. 
We represent scenes as sequences of discrete tokens derived from vector-quantized 3D Gaussian primitives, enabling transformer-based models to incrementally generate, extend, and edit scenes through next-token prediction. 
This formulation combines explicit 3D representations with the compositional inductive biases of autoregressive transformers, providing a complementary alternative to diffusion-based paradigms. 

To sum up, our contributions are: 
\begin{itemize}
\item[(i)] a structured tokenization of Gaussian-based scene representations that enables autoregressive modeling of complex indoor environments
\item[(ii)] an autoregressive transformer framework for incremental 3D Gaussian scene generation, completion, and outpainting.    
\end{itemize}

\section{Related Works}
\paragraph{3D Representations for Generation.}
A critical design choice in 3D generative modeling is the underlying 3D representation. 
Early generative approaches predominantly modeled shapes using discretized or implicit geometric formulations, including voxel-based occupancy grids, signed distance functions, and point clouds \cite{wu_learning_2017, wu_3d_2015, park_deepsdf_2019, chou_diffusion-sdf_2023, shim_diffusion-based_2023, yang_pointflow_2019, achlioptas_learning_2018}.
Meshes constitute another explicit surface representation and have recently been explored for 3D generative modeling \cite{siddiqui_meshgpt_2023, xiang_structured_2025, xiang_native_2025}.
The introduction of Neural Radiance Fields (NeRF) \cite{mildenhall_nerf_2020} enabled high-fidelity 3D generative modeling through continuous volumetric representations \cite{niemeyer_giraffe_2021, muller_diffrf_2023, tang_volumediffusion_2024, poole_dreamfusion_2022}.
However, NeRF-based representations require dense volumetric sampling, resulting in high training and inference costs that limit scalability.
3D Gaussian Splatting \cite{kerbl_3d_2023} has emerged as an efficient and expressive alternative for 3D generative modelling, combining high visual quality with real-time rendering \cite{chen_splatformer_2025, xiang_structured_2025, lan_gaussiananything_2025}. 
However, unconstrained sets of 3D Gaussians lack structural regularity, which makes them difficult to model.
To address this, structured Gaussian parameterizations aligned to spatial grids have been proposed to introduce regularity while preserving rendering efficiency \cite{roessle_l3dg_2024}. 
Furthermore, rather than generating the structured 3D Gaussians directly, we follow the paradigm of latent-space generation \cite{rombach_high-resolution_2022, gao_can3tok_2025}, which has been widely adopted in 3D generative modeling to improve stability, scalability, and sample quality.

\paragraph{Autoregressive 3D Generation.}
Early generative approaches for 3D modeling were largely based on variational autoencoders and generative adversarial networks \cite{wu_3d_2015, wu_learning_2017, achlioptas_learning_2018, yang_pointflow_2019}.
More recently, advances in 3D generative modeling have been driven predominantly by diffusion and flow-matching paradigms, which model complex data distributions through iterative denoising or continuous-time transformations and achieve strong synthesis quality across representations \cite{ho_denoising_2020, lipman_flow_2023, poole_dreamfusion_2022, chou_diffusion-sdf_2023, shim_diffusion-based_2023}.
In parallel, autoregressive modeling, popularized by Transformer-based sequence architectures such as GPT-style models \cite{vaswani_attention_2023, radford_language_2019, brown_language_2020}, has achieved remarkable success in language, vision, and multimodal generation by factorizing joint distributions into a sequence of conditional predictions.
This formulation enables likelihood-based training, progressive generation, and flexible conditioning.
Despite its success in tokenized domains, autoregressive modeling remains comparatively underexplored for 3D generation. 
Existing works primarily focus on sequential modeling of discretized geometric representations, including mesh-based generation \cite{siddiqui_meshgpt_2023, chen_meshanything_2024, hao_meshtron_2024}, hierarchical voxel or octree tokenizations \cite{ibing_octree_2021}, and structured geometric token modeling \cite{chen_mar-3d_2025,han_var-3d_2026,zhang_g3pt_2024}.
In contrast, we explore autoregressive generation over structured Gaussian primitives, enabling scalable modeling of both objects and full scenes while avoiding the iterative sampling procedures characteristic of diffusion-based methods.

\paragraph{From Objects to Scenes.}
On object level, 3D generative modeling has seen remarkable progress in recent years \cite{xiang_structured_2025, lan_gaussiananything_2025, xiang_native_2025, chen_text--3d_2024, tang_dreamgaussian_2024}.
In contrast, full 3D scene generation remains comparatively underexplored. 
Scenes introduce additional challenges, including large spatial scale, long-range spatial dependencies, multi-object compositionality, and the need for coherent layout and physical plausibility. 
Existing approaches address these challenges in different ways. 
Some decompose scene generation into structured layout prediction followed by object retrieval or object-level synthesis \cite{tang_diffuscene_2024, paschalidou_atiss_2021, wang_sceneformer_2021}. 
Others leverage strong priors from pretrained image or video generative models to guide 3D optimization or reconstruction \cite{schneider_worldexplorer_2025, hollein_text2room_2023}, or rely on large-scale foundation models to provide semantic or structural supervision \cite{li_worldgrow_2025, li_dreamscene_2025, zhou_dreamscene360_2024, feng_layoutgpt_2023}. 
A further line of work formulates scene generation as feed-forward reconstruction using depth or multi-view cues combined with Gaussian splatting, focusing on reconstruction rather than unconditional generative modeling \cite{lin_depth_2025, zhang_gs-lrm_2024, jiang_anysplat_2025, charatan_pixelsplat_2024, chen_mvsplat_2025, szymanowicz_flash3d_2025}.
While these strategies demonstrate promising results, they often depend on external priors, multi-stage pipelines, or optimization-heavy procedures. 
In contrast, we aim to learn a unified autoregressive generative model that directly models 3D scenes without relying on pretrained 2D diffusion or video priors, enabling scalable scene synthesis within a single probabilistic framework.


\section{Methodology}
\begin{figure}[t]
  \centering 
  \includegraphics[width=\textwidth, trim={0.5cm, 3cm, 0.5cm, 2cm}, clip]{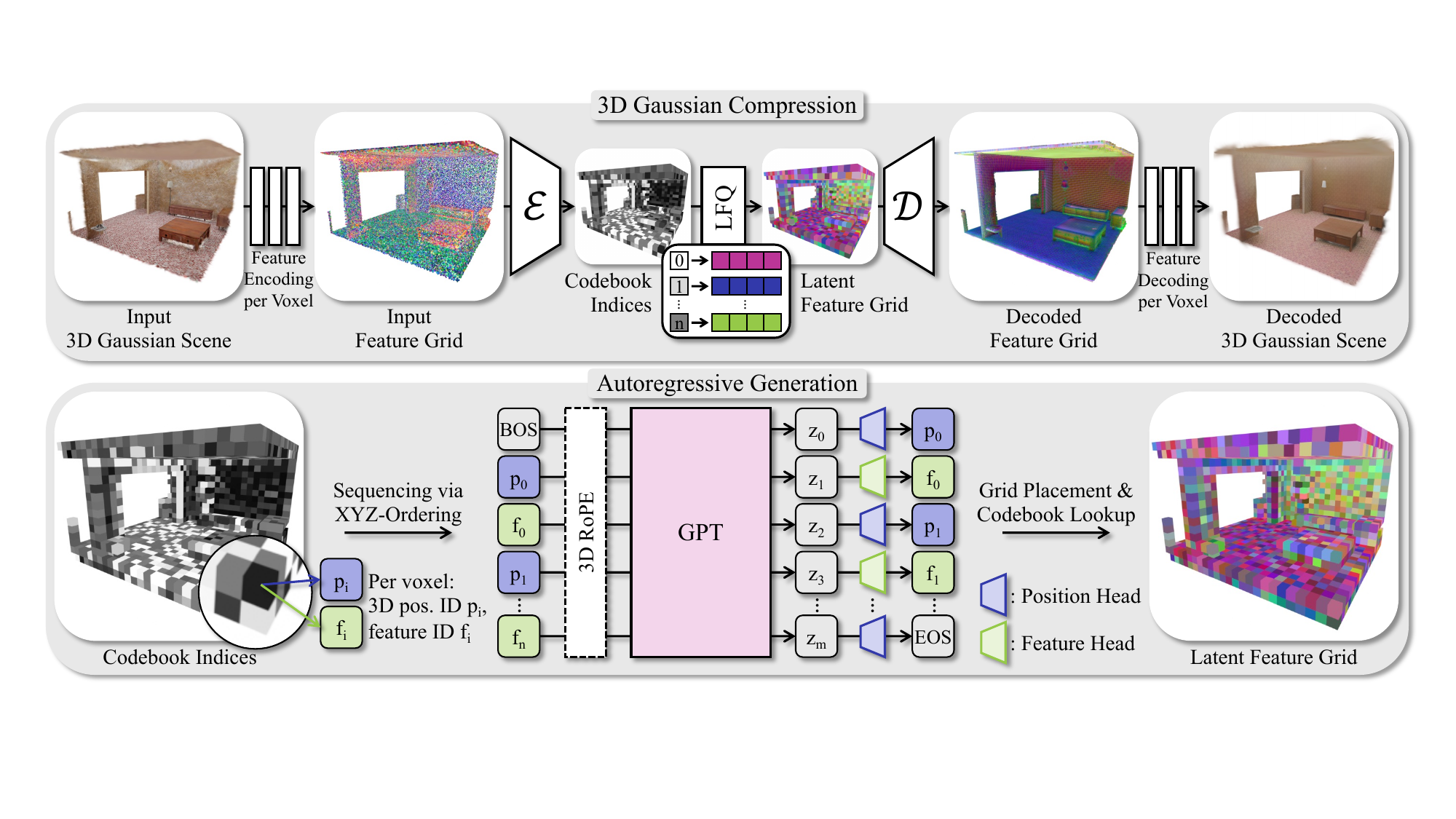} 
    \caption{
    Overview of \textbf{GaussianGPT}.
    A 3D Gaussian scene is subsampled into a sparse voxel grid and encoded into per-voxel features.
    A sparse 3D CNN compresses the grid into discrete codebook indices.
    The resulting latent grid is serialized via $xyz$ ordering and represented as interleaved position tokens $p_i$ and feature tokens $f_i$.
    A causal transformer with 3D RoPE predicts alternating position and feature tokens, which are mapped back to voxel locations and decoded through LFQ to reconstruct the 3D Gaussian scene.
    }
  \label{fig:method}
\end{figure}

Our goal is to synthesize 3D Gaussian scenes using unconditional autoregressive generation with a GPT-style transformer.
Since GPT models operate on discrete token sequences and have limited contextual scope, we first require a compact and discrete representation of 3D scenes.
To this end, we leverage a sparse 3D convolutional autoencoder that maps Gaussian scenes to a discrete latent grid representation and faithfully reconstructs them (\cref{method:ae_part}).
Given this representation, we serialize the latent grids into token sequences and train a causal transformer to model the joint distribution of grid occupancy and features.
An overview of the full pipeline can be seen in \cref{fig:method}.
We detail the autoregressive formulation, architectural design, inference procedure, and the incorporation of 3D spatial priors in \cref{method:gpt_part}.

\subsection{Scene Compression via Sparse 3D Latent Encoding} \label{method:ae_part}
To enable token-based autoregressive modeling, we first transform a continuous 3D Gaussian scene into a compact discrete latent grid representation.
Our compression stage consists of (i) projecting Gaussian primitives into a sparse 3D feature grid, (ii) encoding this grid using a sparse 3D convolutional autoencoder, and (iii) discretizing the latent representation via vector quantization.
The resulting quantized latent grid forms the basis for subsequent autoregressive modeling.

\paragraph{Sparse 3D Feature Grid.}
Following 3D Gaussian Splatting \cite{kerbl_3d_2023}, our input is a scene represented by a set of Gaussian primitives.
Each primitive is defined by a set of continuous attributes: position, opacity, size, rotation, and color.
We define a grid in world coordinates and assign Gaussians to the corresponding voxels based on their positions.
We replace absolute positions with relative offsets from the voxel centers and randomly subsample, when multiple Gaussians are present in a voxel, to obtain our input 3D Gaussian scene.
For each voxel, we use a set of lightweight encoding heads to encode each Gaussian feature, then concatenate the results into a unified vector to obtain a sparse input feature grid.
Symmetrically, after passing through the 3D CNN, a set of per-feature decoding heads converts the predicted per-voxel feature vectors back into individual Gaussian attributes.

\paragraph{Sparse 3D CNN Encoder-Decoder.}
The input feature grid is processed by a sparse 3D convolutional encoder $\mathcal{E}$ and decoder $\mathcal{D}$ following L3DG~\cite{roessle_l3dg_2024}.
The encoder progressively downsamples the grid to a compact latent representation, while the decoder reconstructs voxel-level features.
The convolutional design preserves spatial locality and translation equivariance, yielding structured latent features well suited for subsequent generative modeling.

\paragraph{Vector Quantization.}
In contrast to the L3DG implementation, we adopt lookup-free quantization (LFQ) \cite{yu_language_2024}, which has been shown to improve codebook utilization and quality.
Following the concept of LFQ, the output of our encoder $\mathbf{z}$ directly corresponds to the codebook indices by discretizing to $0$ or $1$ based on the sign.

\paragraph{Training.}
The network is trained on randomly selected images by using a combination of re-rendering, occupancy, and codebook losses.
For the re-rendering losses, $\mathcal{L}_{RGB}$ is an $L_1$ color loss and $\mathcal{L}_{perc}$ is a VGG19-based~\cite{simonyan_very_2015} perceptual loss; both are aggregated over a set of sampled images and poses.
Occupancy predictions in the decoder upsampling layers are guided by a binary cross-entropy loss, $\mathcal{L}_{occ}$, following L3DG.
Lastly, $\mathcal{L}_{LFQ}$ aims to increase the codebook usage by increasing entropy, as introduced in \cite{yu_language_2024}.
We pass the result through an offset softplus function to obtain positive loss values.
\begin{equation}
\mathcal{L} =
\underbrace{ \lambda_{RGB} \mathcal{L}_{RGB} + \lambda_{perc} \mathcal{L}_{perc}}_{\text{re-rendering}}
+
\underbrace{\lambda_{occ} \mathcal{L}_{occ}}_{\text{occupancy}}
+
\underbrace{\lambda_{LFQ} \text{softplus}\left(\mathcal{L}_{LFQ} + 5\right)}_{\text{codebook entropy}} .
\end{equation}

\subsection{Autoregressive Modeling of Latent 3D Grids} \label{method:gpt_part}
Given the quantized latent grid, our goal is to model the joint distribution of its occupancy and features using a sequence-based autoregressive model.
Our first step is therefore to serialize the 3D grid into a 1D token sequence, after which we can train a causal transformer to model the underlying patterns.
We additionally provide priors to the model by separating position and feature vocabularies and injecting 3D positional embeddings.

\paragraph{3D Grid Serialization.}
We linearize the 3D structure into a 1D sequence using a simple and fixed $xyz$ traversal order.
Since $z$ is the least significant dimension, an intuitive understanding of this is that we iterate a scene-height column at every $(x,y)$ position before jumping to the next one.
While this pattern does not preserve 3D locality in the 1D sequence, its advantages lie in its simple, interpretable ordering (see \cref{sec:ar_ablations}).

Given the ordering and relative position indices of the voxels, we construct our sequence by interleaving position and feature tokens.
Since the possible sequence grows cubically with scene size, we limit the required context size by operating our GPT model on chunks rather than entire scenes.
Each voxel is assigned a position index relative to the current grid chunk rather than its absolute position, enabling the model to operate locally and generalize across positions, scenes, and layouts.

\paragraph{Vocabulary Design.}
Although our sequences are processed by a shared transformer backbone, we employ separate vocabularies for position and feature tokens.
Following the regular alternating pattern of position and feature tokens, we alternate between a distinct position and feature head.
The task of the position head is to predict the next occupied voxel index, and the feature head predicts the feature at the preceding position.
This explicit separation in vocabulary and predictions decouples geometric structure from appearance modeling and prevents competition between spatial and semantic features over shared indices.
Incidentally, this design also allows us to freely control the chunk size and the corresponding number of position indices, irrespective of the feature codebook size.

\paragraph{3D Rotary Positional Encoding.}
Transformers, as used in GPT-style models, process inputs as a 1D token sequence and require an explicit notion of position.
In standard applications, positional information is injected along this 1D index, either via learned absolute embeddings or via relative encodings such as rotary positional embeddings (RoPE) \cite{su_roformer_2023}.
However, in our setting, the token sequence is not an inherently one-dimensional signal: it is obtained by serializing a sparse 3D latent grid.
If we were to apply standard 1D positional encoding directly to the serialized order, the model would primarily learn notions of sequence proximity, which can translate poorly to spatial proximity.
In particular, voxels that are close in $(x, y, z)$ may be far apart in the serialization, and vice versa.

To inject an explicit spatial inductive bias, we encode actual voxel coordinates inside the attention mechanism using 3D RoPE.
This way, the attention score becomes a function of the relative spatial offset between the tokens rather than the sequence offset.
The approach follows recent uses of coordinate-aware rotary embeddings in 3D generation~\cite{xiang_native_2025}, enabling the model to reason about the spatial locality independent of the serialization order.

Finally, our sequence alternates between position and feature tokens, both of which correspond to the same 3D location.
In practice, we therefore extend the rotary embeddings with an additional 4th dimension that indicates the token type.
This token-type rotary component helps the transformer to further disentangle geometry from appearance while retaining a single unified attention formulation over the mixed token stream.

\paragraph{Transformer Architecture.}
Our decoder-only causal transformer architecture follows the standard GPT formulation with stacked multi-head self-attention and feed-forward blocks.
Our model is based on GPT-2 \cite{radford_language_2019} and builds on nanochat \cite{nanochat} as its technical backbone.
Compared to the vanilla GPT-2 pipeline, we additionally employ rotary positional embeddings (extended to 3D as described above), query–key normalization, per-layer residual scaling, and the Muon optimizer \cite{henry_query-key_2020, modded_nanogpt_2024, jordan_muon_2024}.
While part of the default configuration of the nanochat backbone, we do not use value embeddings or sliding-window attention \cite{zhou_value_2025, brown_language_2020}.

\paragraph{Training.}
The transformer is trained to model the distribution of serialized scene tokens using a standard autoregressive objective.
Given a tokenized scene sequence $\mathbf{t} = (t_1, \dots, t_T)$, the model predicts each token conditioned on all previous tokens:
\begin{equation}
\mathcal{L}_{CE}
=
- \sum_{i=1}^{T} \log p_\theta(t_i \mid t_{<i}) .
\end{equation}
Training is performed with teacher forcing, where we provide ground-truth tokens as input, and the model learns to predict the next token in the sequence.
Since our representation alternates between position tokens and feature tokens, the cross-entropy loss is computed over the corresponding vocabulary at each step.
Invalid vocabulary entries are masked such that the model only predicts position tokens at position steps and feature tokens at feature steps.

\paragraph{Scene Generation and Completion.}
At inference time, scenes are generated autoregressively by sampling tokens conditioned on previously generated context.
Starting from a begin-of-sequence (BOS) token, the transformer alternates between predicting position tokens and feature tokens until an end-of-sequence (EOS) token is produced.

A key advantage of the autoregressive formulation is that scene completion and unconditional generation are handled by the same mechanism.
Given a partial scene, we simply serialize the observed tokens and use them as a prefix prompt from which the model continues generation.
This allows the transformer to naturally infer missing geometry and appearance while remaining fully consistent with the existing scene context.
The same principle also enables large-scale scene synthesis beyond the fixed training chunk size.
By repeatedly applying completions on a sliding window of previously generated tokens as context, we can continuously outpaint the scene.

Due to the compositional nature of autoregressive generation, we can further limit predictions based on the current sequence.
Specifically, since each position token directly corresponds to a voxel index within the current chunk, we can mask already generated locations, ensuring that the sampled sequence always respects ordering constraints.
Additionally, the sequential nature of our approach enables tree search.
We use this when generating larger scenes by resampling columns without occupancy, thereby creating more occupied and connected scenes.

\section{Experiments} \label{sec:experiments}
We evaluate GaussianGPT on unconditional generation, scene completion, and large scene outpainting.
First, we describe the datasets and experimental setup.
We then present quantitative results on shape generation, scene generation, and scene completion, as well as qualitative results and large-scale outpainting examples.
Lastly, we showcase larger scenes synthesized by our approach and ablate the effect of our design choice for sequence ordering.

\subsection{Datasets}

\paragraph{PhotoShape.}
For object-level experiments, we follow the setup of DiffRF~\cite{muller_diffrf_2023} using PhotoShape~\cite{park_photoshape_2018}.
The dataset contains 15{,}576 chairs normalized to a canonical grid, and associated with 200 rendered views each.

\paragraph{3D-FRONT.}
To obtain high-fidelity Gaussian scenes, we construct a dataset based on 3D-FRONT~\cite{fu_3d-front_2021}.
For each scene, we render multi-view images and initialize Gaussian primitives from depth maps.
These Gaussians are optimized using Scaffold-GS~\cite{lu_scaffold-gs_2023} for $60k$ iterations.
By aligning anchors with voxel centers and restricting each anchor to a single Gaussian, we obtain a high-fidelity Gaussian scene within the input domain of our autoencoder.
After filtering scenes based on spatial extent, our resulting dataset contains 4{,}472 scenes.
To increase data diversity, we additionally apply rotational and reflection-based augmentations, effectively increasing the dataset size by a factor of eight.

\paragraph{Aria Synthetic Environments.}
For large-scale scene modeling, we additionally use Gaussian scenes derived from Aria Synthetic Environments (ASE) \cite{avetisyan_scenescript_2024} optimized as part of SceneSplat++~\cite{ma_scenesplat_2025}.
The dataset contains 25{,}000 indoor scenes with diverse layouts, object arrangements, and appearances.

\subsection{Experimental Setup}

\paragraph{Autoencoder Configuration.}
The scene autoencoder operates on a base voxel size of $0.025\,\mathrm{m}$ and applies three downsampling stages, resulting in a latent grid with a voxel size of $20\,\mathrm{cm}$.
For objects, shapes are discretized on a $128^3$ grid with two downsampling stages to match prior work~\cite{roessle_l3dg_2024}.
We do not model view-dependent appearance for scenes, while objects use first-order spherical harmonics.
Across all experiments, the codebook size is set to 4{,}096.
For more information on the autoencoder hyperparameters, see \cref{app:ae_ablations}.
For the re-rendering losses, we use $\lambda_{RGB}=7.5$ and $\lambda_{perc}=0.3$ with 12 images for scenes, and $\lambda_{RGB}=12.5$ and $\lambda_{perc}=0.1$ with 4 images for objects following \cite{roessle_l3dg_2024}.
$\lambda_{occ}=1.0$ and $\lambda_{LFQ}=0.1$ are the same across all trainings.

\paragraph{Transformer Configuration.}
For scene modeling, we use a GPT-2 medium-sized transformer with a context window of 16{,}384 tokens, which is sufficient to cover a fully occupied chunk.
For object generation, we use a GPT-2 small-sized transformer with a context window of 8{,}192 tokens.
All results are sampled with a temperature of $0.9$ and Nucleus Sampling~\cite{holtzman_curious_2020} with $p=0.9$.

\paragraph{Training Strategy.}
Scene training is performed on spatial chunks with fixed vertical positions in order to align floor heights across scenes and datasets.
During training, we randomly sample chunk locations subject to minimum occupancy thresholds.
For ASE scenes, supervision for autoencoder training is obtained by comparing against rendered ground-truth Gaussians, whereas for 3D-FRONT scenes, we use the previously rendered images.
We limit supervision to image pixels within the corresponding chunk, as defined by the GT geometry.
For object experiments, both the autoencoder and transformer are trained on full shapes without chunking.

\paragraph{Optimization and Compute.}
The autoencoder is trained using Adam~\cite{kingma_adam_2017} with a learning rate of $10^{-4}$, and the transformer uses a combination of AdamW~\cite{loshchilov_decoupled_2019} and Muon~\cite{jordan_muon_2024} with per-module learning rates~\cite{nanochat}.
Both models use cosine learning rate decay to $10\%$ of the initial value.
Autoencoder training uses 4~RTX~A6000 GPUs, with an effective batch size of 8 for scenes and 24 for objects.
Transformer training uses 4 GH200 GPUs with an effective batch size of 64.
We train the autoencoder for around 4 days on scenes and 2 days on PhotoShape. 
GPT training takes around 1 day on scenes and $4.5$ hours on PhotoShape.
We additionally fine-tune both models on 3D-FRONT only, which takes around 1 day for the autoencoder and 10 hours for the GPT model.

\begin{figure}[bt]
  \centering
  \includegraphics[width=\textwidth, trim={2.5cm, 3cm, 0cm, 0cm}, clip]{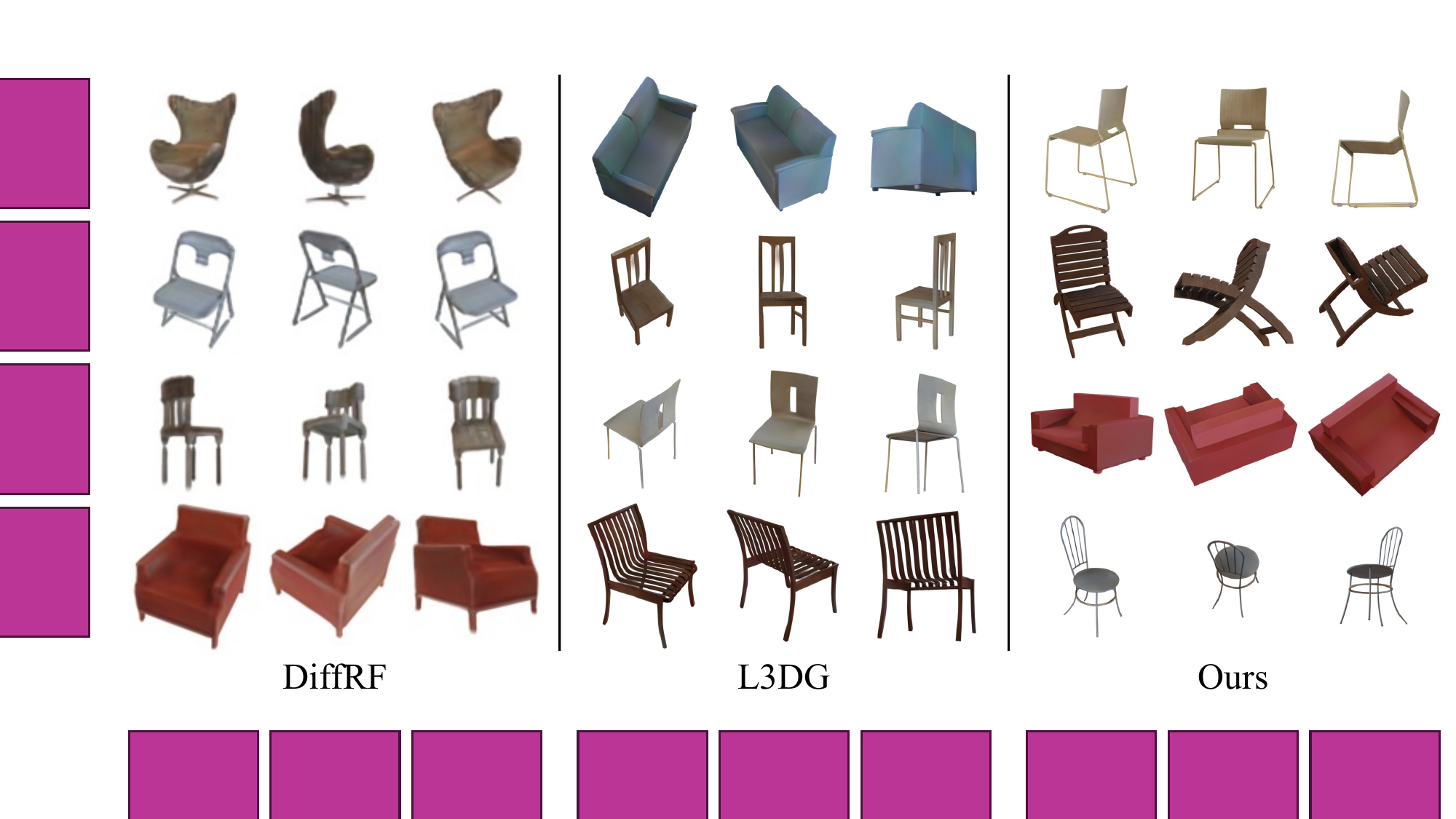} 
    \caption{Qualitative results on unconditional chair generation.
    From left to right: DiffRF~\cite{muller_diffrf_2023}, L3DG~\cite{roessle_l3dg_2024}, and GaussianGPT (ours).
    All models generate high-fidelity shapes; our method produces clean Gaussian allocations with consistent geometric structure.}
  \label{fig:qual_photoshape}
\end{figure}

\subsection{Shape Synthesis}

\begin{table}[tb]
  \caption{Quantitative evaluation of unconditional generation of 1000 shapes after training on the PhotoShape Chairs dataset \cite{park_photoshape_2018}. MMD and KID scores are multiplied by $10^3$.}
  \label{tab:quant_photoshape}
  \centering
  \begin{tabular}{@{}l@{\hspace{12pt}}c@{\hspace{12pt}}c@{\hspace{12pt}}c@{\hspace{12pt}}c@{}}
    \toprule
    Method & FID $\downarrow$ & KID $\downarrow$ & COV $\uparrow$ & MMD $\downarrow$ \\
    \midrule
    $\pi$-GAN \cite{chan_pi-gan_2021}  & 52.71 & 13.64 & 39.92 & 7.387 \\
    EG3D \cite{chan_efficient_2022}    & 16.54 & 8.412 & 47.55 & 5.619 \\
    DiffRF \cite{muller_diffrf_2023}   & 15.95 & 7.935 & 58.93 & 4.416 \\
    L3DG \cite{roessle_l3dg_2024}      & 8.49  & 3.147 & 63.80 & \textbf{4.241} \\
    Ours                               & \textbf{5.68}  & \textbf{1.835} & \textbf{67.40} & 4.278 \\
  \bottomrule
  \end{tabular}
\end{table}

Following the evaluation protocol of L3DG~\cite{roessle_l3dg_2024}, we evaluate GaussianGPT on unconditional chair generation using PhotoShape~\cite{park_photoshape_2018}, 
Rendering quality is measured using Frechet Inception Distance (FID) and Kernel Inception Distance (KID) on $128\times128$ images \cite{heusel_gans_2018, binkowski_demystifying_2021, obukhov2020torchfidelity}.
For geometric evaluation, we compute Coverage (COV) and Minimum Matching Distance (MMD) using Chamfer Distance on $2k$ surface points of meshes extracted from generated Gaussians, following~\cite{achlioptas_learning_2018}.
COV measures sample diversity, while MMD reflects geometric fidelity.

Quantitative results are reported in \cref{tab:quant_photoshape}.
Our autoregressive model achieves the best FID, KID, and COV among all methods, while remaining competitive in MMD, indicating strong alignment with the target distribution in both appearance and geometry.

Qualitative results are shown in \cref{fig:qual_photoshape}.
Both L3DG and GaussianGPT produce high-quality chairs with plausible geometry and appearance.
However, our model tends to avoid noisy outlier primitives, resulting in sharper structures and cleaner renderings while maintaining substantial variation in shape and style.
Overall, these results demonstrate that autoregressive modeling of vector-quantized Gaussian representations can achieve state-of-the-art performance in unconditional 3D shape synthesis.


\subsection{Scene Synthesis}
We evaluate unconditional scene generation and scene completion.
For qualitative comparison, we contrast GaussianGPT with the original L3DG model~\cite{roessle_l3dg_2024}, which is trained to generate full normalized indoor rooms.
To enable a controlled quantitative comparison, we additionally train both GaussianGPT and L3DG on the same chunked 3D-FRONT data under matched spatial settings.
The quantitative results below refer to this matched chunk-level setup.
Qualitative results for the chunked L3DG model are provided in \cref{app:l3dg_chunked_qual}.
\begin{figure}[hbt]
  \centering
  \includegraphics[width=\textwidth, trim={4.2cm, 0cm, 0.9cm, 0cm}, clip]{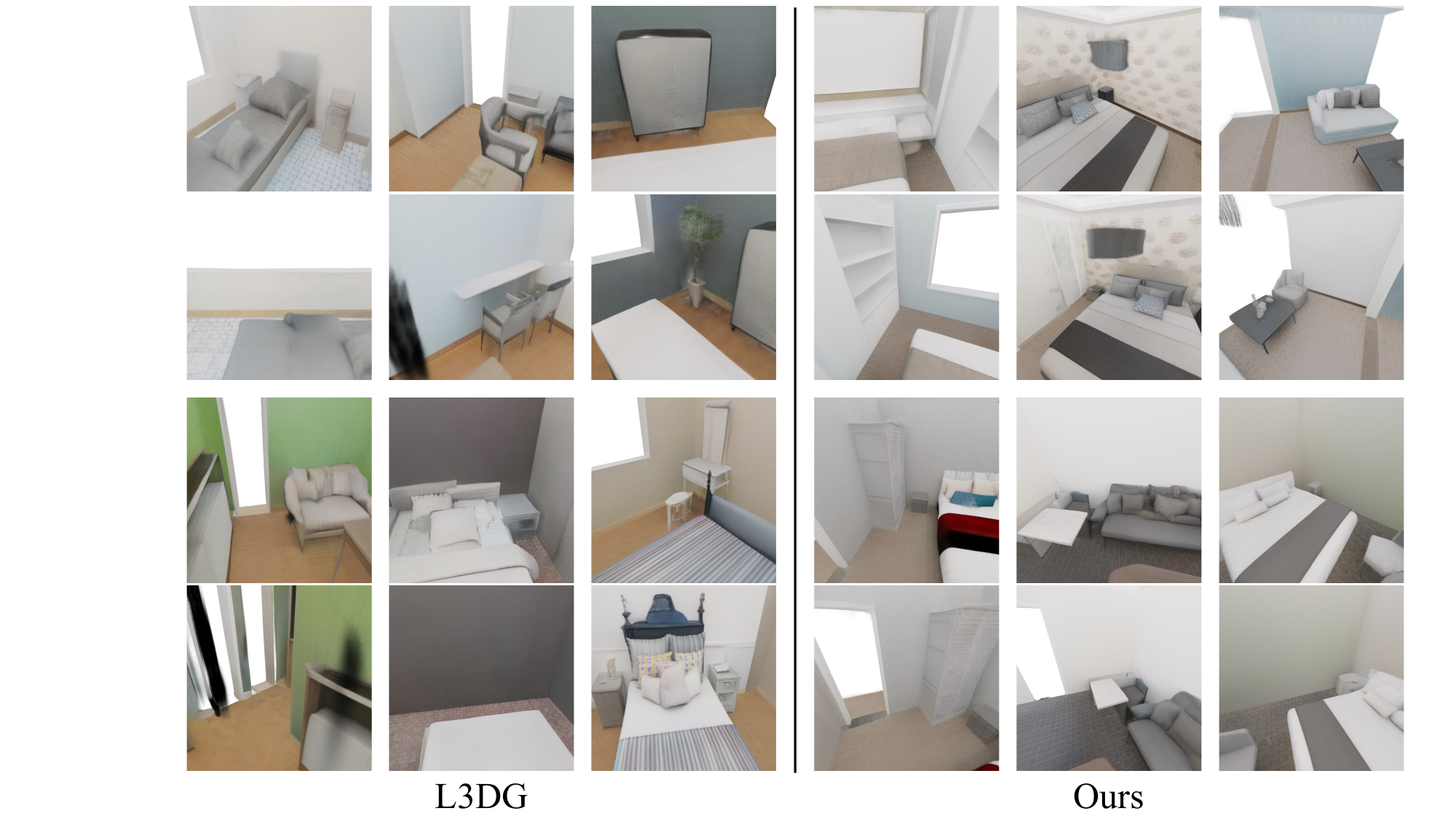} 
  \caption{Qualitative results on unconditional scene generation. We compare the original full-room L3DG~\cite{roessle_l3dg_2024} against GaussianGPT trained only on 3D-FRONT.}
  \label{fig:qual_scenes}
\end{figure}

\paragraph{Evaluation Protocol.}
We evaluate unconditional generation as well as completion from partial context using the metrics in \cref{tab:quant_scenes}.
The given context is a spatially contiguous subset containing the first $25\%$ or $50\%$ in the $x$ dimension.
Appearance quality is measured by FID and KID over rendered views sampled along horizontal orbits around each chunk.
For geometry, we compute Chamfer distance-based COV and MMD from point sets sampled from the Gaussian centers.
Compared to the shape-level evaluation, we raise the resolution to $256 \times 256$ and the number of points to $16k$.
We additionally assess layout quality using FID and KID on top-down depth renderings.
For this, we only keep the bottom $60\%$ of Gaussians per-chunk to remove ceilings.
Completion is evaluated using CLIP similarity between the orbit renderings and their GT equivalents.
We use RePaint \cite{lugmayr_repaint_2022} to evaluate L3DG for completion.

\begin{figure}[htbp]
  \centering
  \includegraphics[width=\textwidth, trim={0cm, 6.5cm, 0cm, 0cm}, clip]{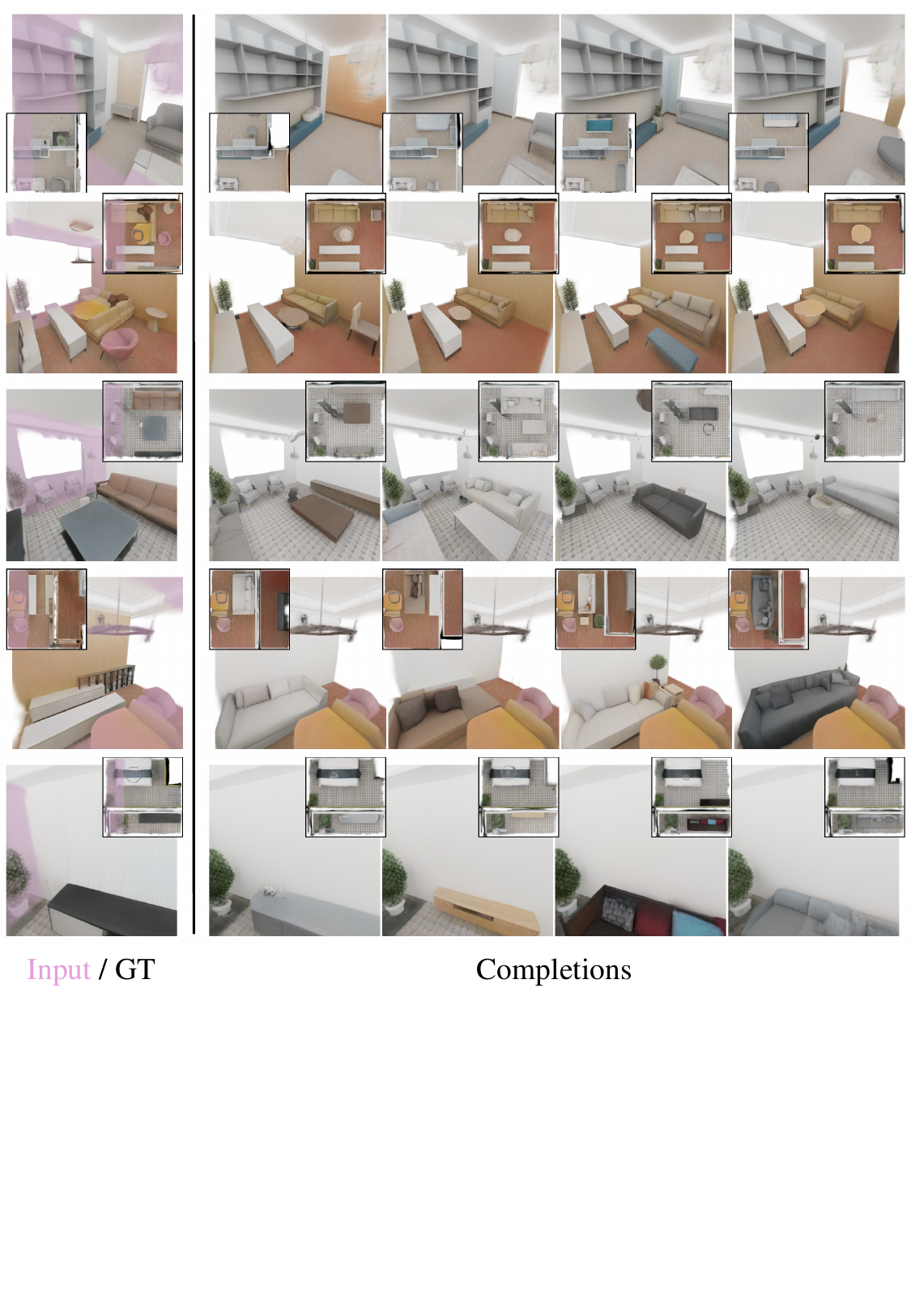}
    \caption{Qualitative results on scene chunk completion. Given one quarter of a validation chunk as context, GaussianGPT produces plausible and diverse completions.}
  \label{fig:completion}
\end{figure}

\paragraph{Qualitative Results.}
As shown in \cref{fig:qual_scenes}, both methods produce coherent indoor layouts with plausible object placement and appearance.
The causal formulation naturally conditions on arbitrary prefixes, enabling scene completion without architectural changes or specialized conditioning mechanisms, as can be seen in \cref{fig:completion}.
We observe that generated completions are both semantically meaningful and diverse across samples, indicating strong modeling of the conditional scene distribution from partial inputs.
\begin{table}[htb]
    \caption{Quantitative evaluation on scene chunks. We compare unconditional generation and completion with $25\%$ and $50\%$ context in terms of appearance, geometry, layout, and completion consistency after training on 3D-FRONT data. Generation is evaluated on $1000$ generated and randomly selected GT chunks, while completion uses $200$ GT chunks with $5$ completions each. GT chunks are selected from held-out scenes.}
  \label{tab:quant_scenes}
  \centering
\begin{tabular}{@{}l@{\hspace{15pt}}cc@{\hspace{3pt}}cc@{\hspace{3pt}}cc@{\hspace{4pt}}c@{}}
  \toprule
  & \multicolumn{2}{c}{Appearance}
  & \multicolumn{2}{c}{Geometry}
  & \multicolumn{2}{c}{Layout}
  & \multicolumn{1}{c}{Completion} \\
\cmidrule(l{1pt}r{1pt}){2-3}
\cmidrule(l{1pt}r{1pt}){4-5}
\cmidrule(l{1pt}r{1pt}){6-7}
\cmidrule(l{1pt}r{0pt}){8-8}
  & FID $\downarrow$
  & KID $\downarrow$
  & COV $\uparrow$
  & MMD $\downarrow$
  & FID $\downarrow$
  & KID $\downarrow$
  & CLIP-Sim.\ $\uparrow$ \\
  \midrule

  \multicolumn{8}{@{}l}{\textit{Generation}} \\
  L3DG & 100.98 & 0.095 & \textbf{0.692} & \textbf{0.096} & 93.84 & 0.092 & -- \\
  Ours & \textbf{94.85} & \textbf{0.084} & 0.548 & 0.118 & \textbf{84.14} & \textbf{0.077} & -- \\

  \midrule
  \multicolumn{8}{@{}l}{\textit{Completion (25\% context)}} \\
  L3DG & 109.35 & 0.100 & \textbf{0.925} & 0.106 & 112.57 & 0.093 & 0.834 \\
  Ours & \textbf{100.06} & \textbf{0.084} & 0.879 & \textbf{0.089} & \textbf{95.60} & \textbf{0.068} & \textbf{0.841} \\

  \midrule
  \multicolumn{8}{@{}l}{\textit{Completion (50\% context)}} \\
  L3DG & 106.54 & 0.096 & \textbf{0.940} & 0.084 & 111.71 & 0.092 & 0.843 \\
  Ours & \textbf{98.13} & \textbf{0.081} & \textbf{0.940} & \textbf{0.057} & \textbf{90.77} & \textbf{0.063} & \textbf{0.851} \\

  \bottomrule
\end{tabular}
\end{table}

\paragraph{Quantitative Results.}
\Cref{tab:quant_scenes} highlights a trade-off for unconditional generation. 
Consistent with the qualitative examples, L3DG produces more varied samples and better matches the ground-truth geometry distribution, while GaussianGPT produces cleaner geometry and stronger appearance and layout metrics.
As the task shifts toward completion and more of the scene is observed, this trade-off increasingly favors our autoregressive formulation.

\begin{figure}[tb]
  \centering
  \includegraphics[width=\textwidth, trim={2cm, 11cm, 2cm, 0cm}, clip]{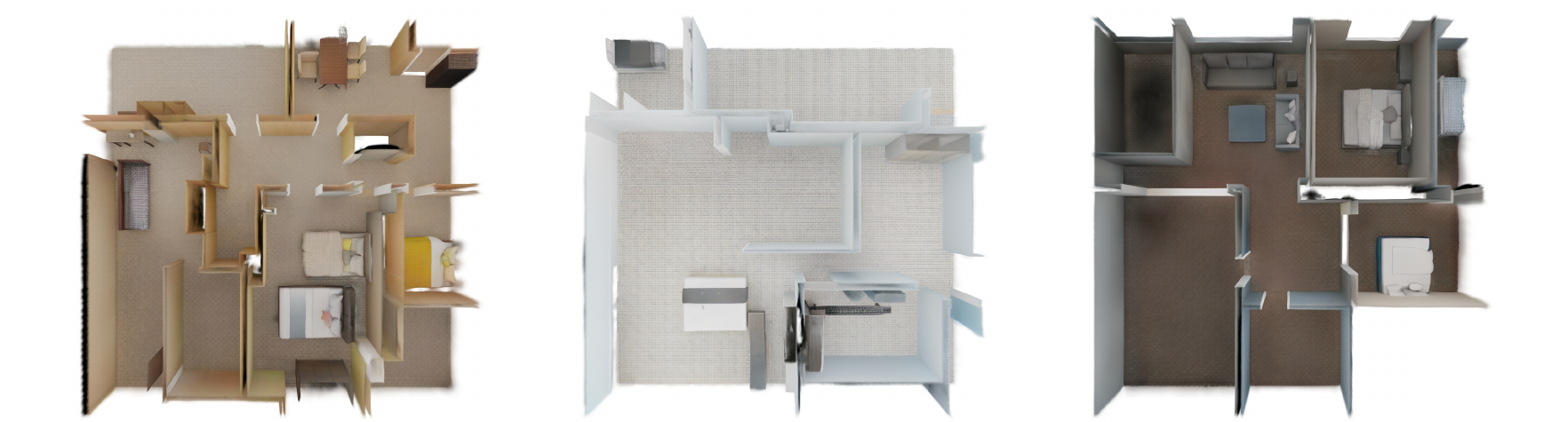} 
  \caption{$12\,\mathrm{m} \times 12\,\mathrm{m}$ scene synthesis via autoregressive outpainting.
GaussianGPT sequentially generates and appends latent grid columns, enabling scenes to grow beyond the training horizon.}
\vspace{-0.2cm}
  \label{fig:big_scenes}
\end{figure}

\paragraph{Large Scene Generation.}
As shown in \cref{fig:big_scenes}, our model can extend scenes beyond the fixed training horizon.
We iteratively generate and append latent-grid columns using previously generated regions as context.
To promote occupancy across larger scene generation, we leverage the step-by-step generative process to introduce a backtracking resampling strategy.
If a column without any geometry is predicted, we move back to the last prior token and attempt sampling again up to 5 times.
GaussianGPT produces coherent and stylistically consistent extended scenes over substantial spatial ranges, although quality gradually degrades with distance from the initial context, as analyzed in \cref{app:largescenegen}.
These results highlight the flexibility of autoregressive modeling for open-ended 3D scene synthesis, where scene size is not predetermined, and generation can naturally scale with context.

\subsection{Autoregressive Modeling Ablations} \label{sec:ar_ablations}
We ablate key design choices of our autoregressive model on the combined ASE and 3D-FRONT training setup.
Specifically, we study how the 3D latent grid is serialized into a token sequence and how spatial and feature information are represented within the transformer.
For all variants, we keep the remaining architecture and training configuration fixed and report average training and validation cross-entropy over epochs 70--74.

\paragraph{Serialization Strategies.}
\Cref{tab:quant_orderingablation} compares different 3D-to-1D serializations.
Besides our column-wise $xyz$ traversal, we evaluate Z-order and Hilbert space-filling curves and their transposed variants following Point Transformer V3 \cite{wu_point_2024}.
Despite stronger locality guarantees, these orderings do not improve performance.
The simple $xyz$ traversal performs best overall, while transposed Z-order achieves nearly identical validation performance.
This suggests that 3D RoPE already provides sufficient spatial information, reducing the importance of locality in the serialized sequence.

\paragraph{Core Model Components.}
\Cref{tab:quant_componentablation} evaluates separate vocabularies for position and feature tokens, as well as 3D RoPE.
A shared vocabulary increases validation cross-entropy, while learned positional embeddings or 1D RoPE degrade both training and validation performance.
Together, these results show that separating spatial and feature prediction and explicitly encoding 3D coordinates provide useful inductive biases for autoregressive scene modeling.

\begin{table}[htb]
\centering
\begin{minipage}[t]{0.48\linewidth}
  \vspace{0pt}
  \centering
  \caption{Ablation of 3D-to-1D serialization strategies. 
  }
  \label{tab:quant_orderingablation}
  \begin{tabular}{l@{\hspace{4pt}}c@{\hspace{4pt}}c}
    \toprule
    Ordering & Train CE $\downarrow$ & Val CE $\downarrow$ \\
    \midrule
    Z-order            & 2.203 & 2.432 \\
    Trans. Z-order     & 2.204 & 2.423 \\
    Hilbert            & 2.347 & 2.439 \\
    Trans. Hilbert     & 2.334 & 2.434 \\
    $xyz$ (ours)       & \textbf{2.151} & \textbf{2.421} \\
    \bottomrule
  \end{tabular}
\end{minipage}
\hfill
\begin{minipage}[t]{0.48\linewidth}
  \vspace{0pt}
  \centering
  \caption{Ablation of the core design decisions: separate vocabulary and 3D RoPE.}
  \label{tab:quant_componentablation}
  \begin{tabular}{l@{\hspace{4pt}}c@{\hspace{4pt}}c}
    \toprule
    Setup  & Train CE $\downarrow$ & Val CE $\downarrow$ \\
    \midrule
    Ours                & \textbf{2.151} & \textbf{2.421} \\
    \midrule
    w/ Shared Vocab.    & 2.157 & 2.449 \\
    w/ Learned PE       & 2.210 & 2.461 \\
    w/ 1D RoPE          & 2.227 & 2.446 \\
    \bottomrule
  \end{tabular}
\end{minipage}
\end{table}

\section{Conclusion}
We introduced \textbf{GaussianGPT}, a fully autoregressive framework for 3D scene generation and completion operating directly on vector-quantized Gaussian representations.
By combining sparse 3D latent compression with structured tokenization and 3D-aware transformer modeling, we show that sequential next-token prediction is a viable and complementary alternative to diffusion-based approaches for structured 3D synthesis.
Compared to diffusion-based methods, our formulation enables several new capabilities.
Scenes can be generated incrementally as a sequence of discrete spatial decisions, allowing explicit control over sampling, causal reasoning over partially generated geometry, and seamless support for completion and outpainting using the same model.

We demonstrate improved visual quality in 3D shape synthesis and scene-level generation and completion, with stronger perceptual and layout quality while retaining a trade-off in geometric diversity.
More broadly, our work highlights the potential of treating 3D scenes as structured token sequences, opening new opportunities for controllable and compositional 3D generation.
Future work includes extending the framework to real-world data, where uncertainty-aware modeling becomes essential, and autoregressive approaches are particularly well-suited.
We also imagine other promising avenues for mitigating large-scene degradation, improving long-horizon stability, and extending the generation context beyond fixed spatial chunks and orderings.

\clearpage
\section*{Acknowledgements}
Nicolas von Lützow is supported by the MDSI focus topic \emph{Understanding Existing Structures in Building Planning} (USP), and Katharina Schmid is an MDSI doctoral fellow.
This work was further supported by the ERC Consolidator Grant \emph{Gen3D} (101171131), and by compute resources from the Jülich Supercomputing Centre under project \emph{MeshFoundation}.
We thank Angela Dai for the video voice-over.

%
%
\bibliographystyle{splncs04}
\bibliography{main}

\clearpage
\begin{center}
{\normalfont\Large\bfseries\boldmath Appendix}
\end{center}

\appendix

\section{Real-World Results} \label{app:realworld}
To assess the feasibility of our approach on real-world data, we fine-tune our models on ScanNet++~v2~\cite{yeshwanth_scannet_2023}.
Compared to the synthetic datasets used in the main paper, this setting introduces several additional challenges: higher visual and geometric complexity, greater scene variability, and the absence of globally consistent axis-alignment for walls and floors.
Moreover, the available dataset is significantly smaller.
We use Gaussians optimized as part of SceneSplat++~\cite{ma_scenesplat_2025}, obtaining $895$ scenes in total, of which $90\%$ are used for training.
Even with the same $8\times$ rotation and flipping augmentation as in the main experiments, the effective training set remains limited.

\paragraph{Training Details.}
We obtain DSLR images from the official ScanNet++ v2 release and fine-tune the autoencoder using chunked training at half resolution for $100$ epochs (approximately one day of training).
The GPT model is then fine-tuned on the resulting latent grids for $250$ epochs, which takes around $7.5$ hours.

\begin{figure}[hbt]
  \centering
  \includegraphics[width=\textwidth, trim={0cm, 11cm, 0cm, 0cm}, clip]{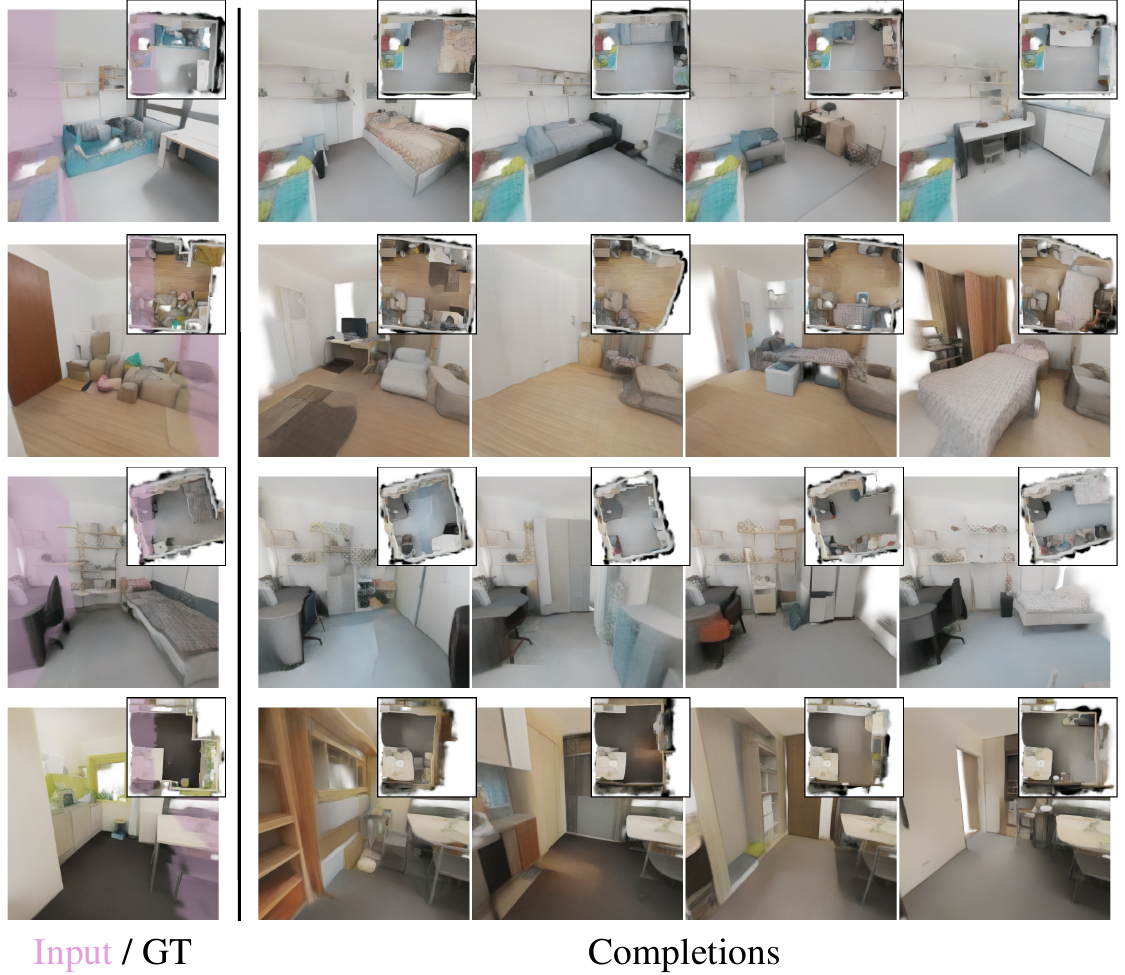}  
  \caption{Qualitative results on scene chunk completion on ScanNet++~v2~\cite{yeshwanth_scannet_2023}. We sample completions with temperature $0.6$, and Nucleus Sampling \cite{holtzman_curious_2020} with $p=0.7$, yielding varied and feasible completions.}
  \label{fig:real_world_completions}
\end{figure}

\paragraph{Completion Results.}
We evaluate the model on scene chunk completion using the same setup as in the main paper and show results in \cref{fig:real_world_completions}.
Despite the small dataset, the model retains useful priors learned from synthetic data and produces diverse and plausible completions in terms of geometry, texture, and semantic layout.

\paragraph{Limitations.}
However, we observe that the fidelity of the autoencoder becomes a limiting factor in the real-world setting.
In particular, high-frequency details are often not fully reconstructed, leading to noisier geometry and Gaussian predictions.
Furthermore, real-world scans inherently contain missing or unobserved regions that cannot be faithfully modeled by the current pipeline.

\paragraph{Future Work.}
Improving reconstruction fidelity and explicitly modeling uncertainty in partially observed environments remain important directions for future work.
Autoregressive formulations are particularly well suited for this setting, as their compositional sequence of small, discrete decisions allows uncertain regions to be masked while still enabling probabilistic reasoning based on the available conditioning signals.
This could enable training on partial real-world scans while internally modeling and generating complete scenes, potentially unlocking further advantages of autoregressive approaches in real-world environments.

\clearpage
\section{Inference Efficiency Analysis} \label{app:inference_efficiency}
We benchmark inference efficiency in \cref{tab:inference_efficiency} by measuring per-chunk wall-clock time and peak GPU memory on a single A100 GPU.
For both unconditional generation and completion from partial context, we report mean runtime and standard deviation across 100 samples.
To isolate the cost of the respective base sampling procedures, we disable resampling for our method and do not use backward repainting steps for completion with L3DG.

\newcommand{\std}[1]{\,{\scriptsize\(\pm\) #1}}
\begin{table}[htb] 
  \centering
\caption{Inference efficiency analysis on a single A100 GPU. We report mean wall-clock time with standard deviation and mean peak GPU memory over 100 samples.}
  \label{tab:inference_efficiency} 
  \small
  \begin{tabular}{c@{\hspace{15pt}}c@{\hspace{15pt}}r@{}l@{\hspace{15pt}}r}
    \toprule
    Task & Approach & \multicolumn{2}{c}{Wall-clock} & Memory \\
    \midrule
    \multirow{2}{*}{Unconditional}
      & L3DG & \textbf{23.8\,s} & \std{\textbf{0.07\,s}} & \textbf{1.55\,GB} \\
      & Ours & 78.1\,s & \std{15.8\,s} & 5.04\,GB \\
    \midrule
    \multirow{2}{*}{25\% Context}
      & L3DG & \textbf{24.0\,s} & \std{\textbf{0.18\,s}} & \textbf{1.59\,GB} \\
      & Ours & 54.5\,s & \std{13.8\,s} & 5.24\,GB \\
    \midrule
    \multirow{2}{*}{50\% Context}
      & L3DG & \textbf{24.1\,s} & \std{\textbf{0.14\,s}} & \textbf{1.60\,GB} \\
      & Ours & 38.9\,s & \std{12.4\,s} & 5.46\,GB \\
    \bottomrule
  \end{tabular}
\end{table}

Autoregressive sampling is slower than the diffusion-based L3DG baseline for unconditional generation, even when considering the more compressed latent space.
This difference primarily reflects the sequential sampling of next-token in the sequence, while diffusion models are able to update the full representation in parallel at each denoising step.
However, when considering completion from partial context, we can see that sampling time drastically reduces for ours, while there is no change for the diffusion baseline that continues to compute equivalent update steps.

For our autoregressive model, memory is dominated by the key-value cache sized for the full context window, since the final sequence length is unknown before sampling. 
Introducing a prefix constitutes a small bump during prefill, but generally, peak memory remains nearly constant across unconditional generation and completion,
It should therefore be interpreted as a conservative implementation-level bound rather than the memory required by the realized sequence. 
In contrast, L3DG uses a fixed dense grid during generation, requiring approximately $1.25\, \mathrm{GB}$. 
Its higher reported peak memory is incurred during decoding.

From an efficiency perspective, this positions AR sampling as a promising alternative for scene completion or inpainting settings, where large parts of the scene are already known.
Additionally, there remains clear potential to further accelerate autoregressive sampling:
Our method relies on long sequences to represent scenes, and as such, decode-optimized kernels could significantly speed up inference by parallelizing key/value cache accesses \cite{dao_flash_decoding_2023}.
Another option is speculative decoding, in which a smaller draft model proposes several future tokens that are verified in parallel by the full model, thereby reducing the number of expensive autoregressive decoding steps \cite{leviathan_fast_2023, chen_accelerating_2023}.
Similarly, the strict column-wise structure of our predictions may be well suited to auxiliary decoding heads that propose multiple future tokens per step for parallel verification \cite{cai_medusa_2024}.
Moreover, the spatial structure of the sequence could enable sparse or locality-aware attention patterns, reducing attention cost by focusing computation on nearby or otherwise relevant regions while retaining occasional global interactions.
More fundamentally, hierarchical or adaptive tokenizations, as well as stronger compression more generally, could reduce the number of tokens required to represent a scene, thereby simplifying the modeling task and accelerating inference.

\section{Large Scene Generation} \label{app:largescenegen}
\begin{figure}[bht]
  \centering
  \includegraphics[width=\textwidth, trim={5cm, 3cm, 5cm, 0cm}, clip]{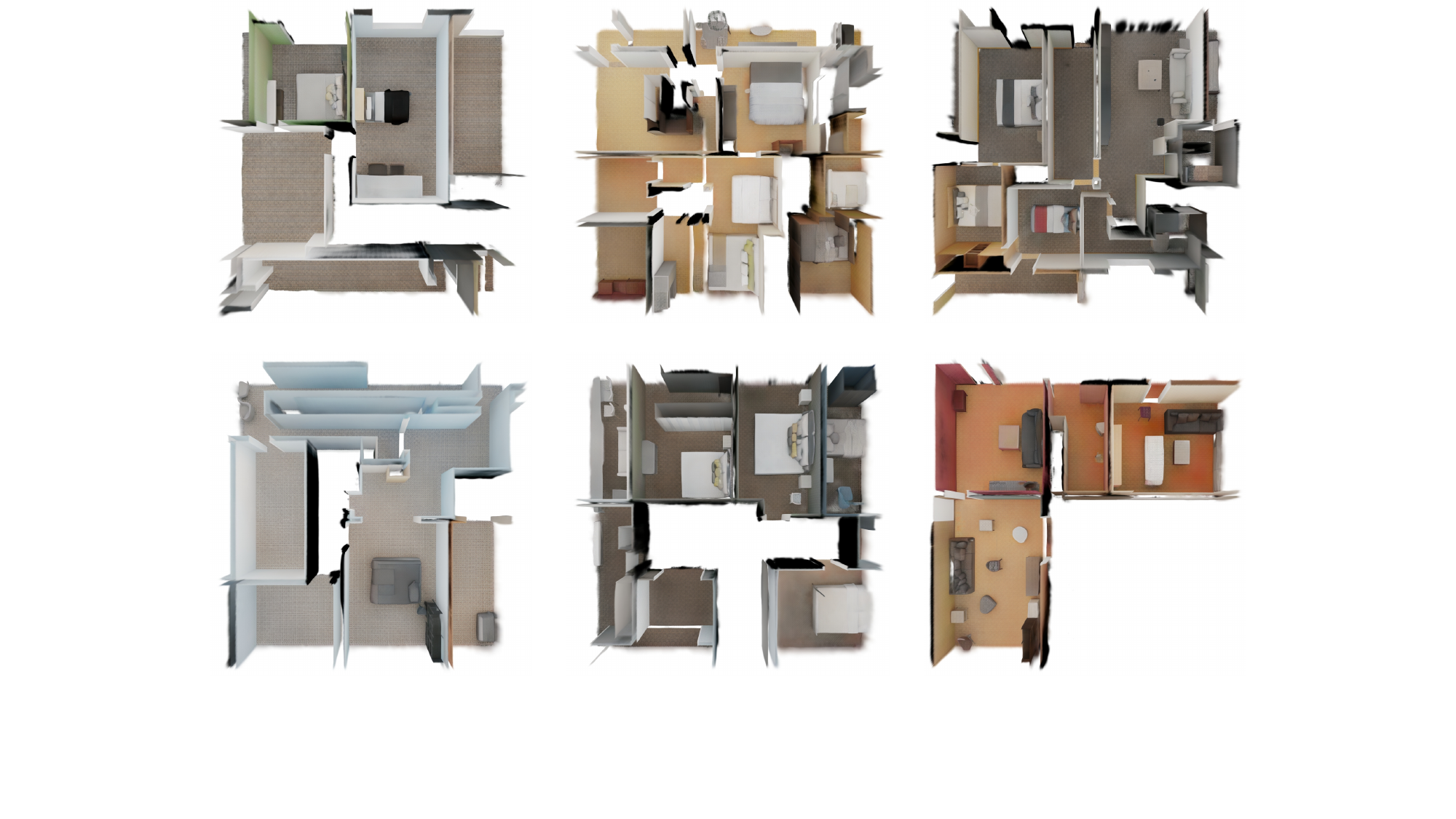}  
  \caption{Additional examples of $12\,\mathrm{m} \times 12\,\mathrm{m}$ scene synthesis.}
  \label{fig:big_scenes_app}
\end{figure}

For large-scale scene generation, we extend scenes via autoregressive outpainting using a sliding-window local context.
When predicting a new column, the model conditions on previously generated columns within the local chunk and predicts the next column within this local coordinate frame.
The used $xyz$ ordering can be naturally extended to larger scenes while keeping the relative ordering consistent across the global sequence and local chunks.
Additional results are shown in \cref{fig:big_scenes_app}.

\paragraph{Sampling Efficiency.}
Scene generation proceeds in three stages.
First, an initial contiguous context block is produced via single-chunk sampling.
Second, a shifted bootstrap stage expands this seed along the $x$-direction by sliding the context window and generating a new slice of 20 columns per step.
Finally, the main outpainting loop grows the scene along the $y$-direction until the target size is reached.
Bootstrapping reduces the number of sampling loop calls by generating multiple columns per call and enables efficient reuse of the transformer KV-cache.
While the sparse nature of our representation leads to large variance in generation times, on average, synthesizing a $12\,\mathrm{m} \times 12\,\mathrm{m}$ scene takes approximately $6{,}000\,\mathrm{s}$ on a GH200 GPU.
Generating a single $4\,\mathrm{m} \times 4\,\mathrm{m}$ chunk, however, only takes roughly $80\,\mathrm{s}$. 
This corresponds to about 5 columns/s for fully KV-cached single-chunk generation, dropping significantly to only around $0.6$ columns/s for large-scene generation where caching is not possible.
Without resampling, generation is approximately twice as fast.

\paragraph{Chunk Window Positioning.}
Another practical consideration is the placement of the prediction target within the sliding window.
During training, the model learns to predict columns at arbitrary positions inside the local context.
At inference time, we exploit this flexibility to balance the amount of history behind the prediction with the available local context.
In particular, we generate up to 5 columns at a time, starting from $y$ at the chunk center.
%
%
We also aim to predict around $x=5$, allowing each local window to resemble the distribution of training chunks due to the unknown section.
Together, these choices yield a stable, well-conditioned generation frontier during large scene synthesis.

\begin{figure}[htb]
  \centering
  \includegraphics[width=\textwidth]{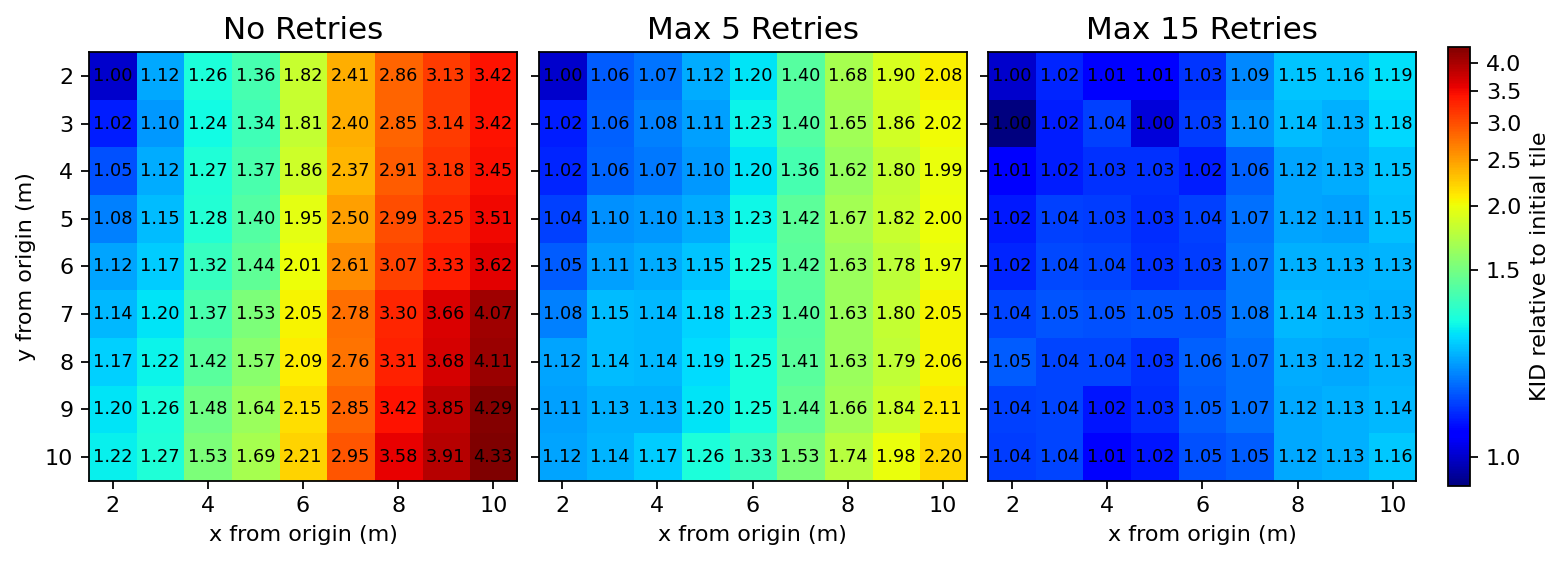}  
    \caption{%
        Large-scene degradation analysis. KID per $4\,\mathrm{m} \times 4\,\mathrm{m}$ tile
      across a $12\,\mathrm{m} \times 12\,\mathrm{m}$ scene, normalized to the top-left initial tile
      ($1.0=$ no degradation; warmer $=$ worse). Retries suppress spatial
      degradation. Chunks are labeled by their center, and the color scale is shared across grids.%
    }
  \label{fig:degradation}
\end{figure}

\paragraph{Resampling and degradation.}
Training chunks can contain relatively sparse geometry, which may lead to empty predictions when generating larger scenes. 
To encourage meaningful occupancy, we apply a simple resampling strategy: whenever a predicted column is empty, we retry generation up to a fixed number of times.
We analyze quality degradation and the effect of this strategy in \cref{fig:degradation}. 
As expected for autoregressive sampling, quality gradually deteriorates with increasing distance from the initial chunk as prediction errors accumulate. 
Resampling helps mitigate this effect by preventing geometry from drifting into empty or out-of-distribution regions, thereby delaying the onset of degradation. 
We evaluate the performance with maximum retires set to 0, 5, and 15.
While 15 retries further improves quality and delays degradation, we use 5 retries by default as a more computationally efficient trade-off.

Interestingly, degradation is substantially stronger along the $x$ direction than along $y$. 
We attribute this asymmetry to the $xyz$ serialization order: extending a scene along $y$ remains relatively local in the token sequence, whereas extending it along $x$ requires larger spatial jumps.
We additionally experimented with a hard constraint that restricts predictions to the current or next column, effectively enforcing full occupancy. 
This performed poorly in practice, likely because it prevents the model from predicting empty regions even when they are substantially more probable than occupied ones.

\section{Additional Training Details} \label{app:trainingdetails}
\paragraph{Chunk Sampling.}
Both the autoencoder and the GPT model are trained by sampling spatial chunks from the available scenes.
For each training step, up to $10$ candidate chunks are proposed until one satisfies the minimum occupancy threshold of $0.2$ for autoencoder training and $0.3$ for GPT training, since the latent grid is coarser.
If no candidate meets the requirement, the chunk with the highest occupancy is selected instead.

\paragraph{Camera Sampling.}
For autoencoder training, we additionally sample cameras conditioned on the selected chunk.
Views that meaningfully observe the chunk are prioritized.
For 3D-FRONT~\cite{fu_3d-front_2021}, we can determine the relevant image area precisely by comparing depth maps with the chunk boundaries.
We therefore sample $8\times$ the required number of images, score them based on the amount of visible chunk area, and sample proportionally to this score.
This approach favors informative views while still allowing the use of all available camera viewpoints.
For ASE~\cite{avetisyan_scenescript_2024, ma_scenesplat_2025}, we use a simpler heuristic based on the projected chunk bounding box in screen space.
Cameras containing at least $40\%$ chunk coverage are preferred, falling back to cameras with any overlap if necessary.

\paragraph{Encoder and Decoder Heads.}
In our autoencoder, Gaussian attributes are encoded into and predicted from feature grids using dedicated encoder and decoder heads.
Each encoder head consists of a linear layer that expands the feature dimensionality by a factor of $16$, followed by a residual MLP block that preserves dimensionality.
The resulting feature embeddings are concatenated and passed to the encoder network.
After decoding, the reconstructed feature vectors are processed by individual decoder heads.
Each decoder head consists of two residual MLP blocks with $64$ channels, followed by a final linear projection to the attribute's dimensionality.
Separate heads predict scales, opacities, quaternions, color, coordinate offsets, and spherical harmonics where applicable.
To impose a controlled prior at the beginning of training, the final projection layers are initialized with zero weights and constant biases chosen to ensure reasonable initial visibility.

\paragraph{Feature Representations.}
Each Gaussian attribute can be parameterized in different representations and spaces.
For autoencoder training, we represent scales in world-space size using a softplus activation, opacities in logit space clamped to $[-10,10]$, colors clamped to $[0,1]$, standardized quaternions, and offsets as unbounded world-space values.
We additionally scale features to similar magnitudes to stabilize training.

\section{GPT Training Hyperparameters}
%
The used nanochat \cite{nanochat} backend is tuned for fast convergence on language tasks and contains per-module optimization hyperparameters.
Our implementation is based on commit \textit{e527521}.
We find their hyperparameters to work well in our setting and converge quickly and stably.
\cref{tab:gpt_hps} shows the per-module optimizer, learning rate, and additional parameters.
\begin{table}[htb]
  \caption{Learning Rates and Optimizer Parameters used in our GPT model training. 
  $\dag$ Learning Rates are scaled by embedding dimension; the given values are exact for object training. For scenes, they are divided by $\sqrt{1024 / 768} \approx 1.155$.
  }
  \label{tab:gpt_hps}
  \centering
  \begin{tabular}{@{}l@{\hspace{12pt}}c@{\hspace{12pt}}c@{\hspace{12pt}}c@{}}
    \toprule
    Module & Optimizer & Learning Rate & Optimizer Parameters \\
    \midrule
    Output Heads       & AdamW & 0.004 \dag & $\beta_1 = 0.8, \beta_2=0.95$  \\ 
    Embeddings         & AdamW & 0.2 \dag   & $\beta_1 = 0.8, \beta_2=0.95$  \\ 
    Residual Weights   & AdamW & 0.005      & $\beta_1 = 0.8, \beta_2=0.95$  \\ 
    Input Skip Weights & AdamW & 0.5        & $\beta_1 = 0.96, \beta_2=0.95$ \\
    Others             & Muon  & 0.02       & $\beta_2 = 0.95, n=5 $ \\
  \bottomrule
  \end{tabular}
\end{table}

\paragraph{Optimizers.}
All modules optimized using AdamW \cite{loshchilov_decoupled_2019} additionally use $\epsilon = 10^{-10}$ for stability and no weight decay.
For optimization using Muon \cite{jordan_muon_2024}, $\beta_2$ refers to the weight used in variance reduction and $n$ to the number of Newton-Schulz iterations \cite{bernstein_modular_2024, higham_functions_2008, bjorck_iterative_1971, kovarik_iterative_1970}.
The used Nesterov momentum \cite{sutskever_importance_2013} increases linearly from $0.85$ to $0.95$ over the first 300 iterations.
Additionally, Muon uses a weight decay of $0.025$ for scenes and $0.1$ for objects, scaled linearly to $0$ over the duration of training.

\paragraph{Weight Initialization.}
Token embeddings are initialized from $\mathcal{N}(0,1)$ and the output heads from $\mathcal{N}(0,10^{-3})$.
All attention projections and MLP input layers are initialized uniformly between $\pm \sqrt{3/d}$, where $d$ is the embedding dimension. 
Attention and MLP output projections are initialized to zero.
Residual weights are initialized to $1.0$, and input skip weights to $0.1$.

\section{Autoencoder Ablations}\label{app:ae_ablations}
Our autoregressive model operates on discrete latent representations produced by the autoencoder.
We therefore study the reconstruction quality and efficiency trade-offs induced by the autoencoder architecture and tokenizer in \cref{tab:ae_ablations}.
Specifically, we evaluate reconstruction fidelity on held-out 3D-FRONT scenes by comparing renderings of reconstructed chunks against renderings of the ground-truth meshes, reporting PSNR, SSIM, and LPIPS to measure pixel-wise fidelity, structural similarity, and perceptual similarity, respectively.
For the codebook-size ablations, we additionally report code usage, defined as the percentage of codes used at least once during validation.

\paragraph{Downsampling.}
The main trade-off is governed by the number of downsampling stages.
Reducing downsampling improves reconstruction quality, but substantially increases the number of latent tokens that must subsequently be modeled autoregressively.
On average, the $1{\times}$, $2{\times}$, and $3{\times}$ configurations produce approximately $53k$, $14k$, and $3.2k$ latent voxels per scene, respectively.
While the $1{\times}$ and $2{\times}$ variants achieve higher reconstruction fidelity after 20 epochs, their substantially longer sequences make them infeasible for downstream autoregressive modeling.

\paragraph{Codebook Size and Type.}
We further compare different codebook sizes and quantization schemes.
Varying the codebook size produces only minor differences in reconstruction quality.
Increasing the codebook size to $16384$ slightly improves reconstruction quality, but code usage drops substantially to $86.4\%$.
This indicates that the larger codebook is not utilized efficiently, making the limited gain in reconstruction quality difficult to justify downstream, where well-behaved codebook usage is desired.
Traditional vector quantization with a codebook \cite{oord_neural_2018} provides slightly stronger reconstruction performance early in training, although this gap narrows with continued optimization.
However, it incurs approximately one-third longer training time due to the required nearest-neighbor searches.
Since both the autoencoder and tokenizer benefit substantially from extended training, we instead use LFQ for improved computational efficiency.
This is reflected in the final model, whose reconstruction metrics improve considerably beyond the 20-epoch results.

\begin{table}[htb]
  \centering
  \caption{Autoencoder ablations on 3D-FRONT. We evaluate reconstruction quality on held-out scene chunks after 20 training epochs. In the first block, we vary the number of downsampling stages and report the average number of latent voxels per chunk. In the second block, we vary codebook size and quantization scheme, reporting the fraction of the codebook used during validation.}
  \label{tab:ae_ablations}
  \small
  \begin{tabular}{l@{\hspace{10pt}}c@{\hspace{10pt}}c@{\hspace{10pt}}c@{\hspace{10pt}}r}
    \toprule
    Configuration & PSNR $\uparrow$ & SSIM $\uparrow$ & LPIPS $\downarrow$ & Latents / Usage \\
    \midrule
    $1\times$ Downsampling        & \textbf{24.34} & \textbf{0.873} & \textbf{0.187} & 53k Latents \\
    $2\times$ Downsampling        & 23.73 & 0.869 & 0.190 & 14k Latents \\
    $3\times$ Downsampling (ours) & 21.11 & 0.847 & 0.246 & \textbf{3.2k Latents} \\
    \midrule
    1024 CB                       & 20.93 & 0.846 & 0.250 & \textbf{100\% Usage} \\
    4096 CB (ours)                & 21.11 & 0.847 & 0.246 & 99.2\% Usage \\
    16384 CB                      & 21.37 & 0.848 & 0.245 & 86.4\% Usage \\
    VQ w/ Codebook                & \textbf{21.59} & \textbf{0.850} & \textbf{0.226} & 98.4\% Usage \\
    \midrule
    Full Training                 & 25.04 & 0.898 & 0.134 & \\
    \bottomrule
  \end{tabular}
\end{table}

\clearpage
\section{Chunked L3DG Results} \label{app:l3dg_chunked_qual}
\begin{figure}[htb]
  \centering 
  \includegraphics[width=\textwidth, trim={0.5cm, 19cm, 0.5cm, 0.2cm}, clip]{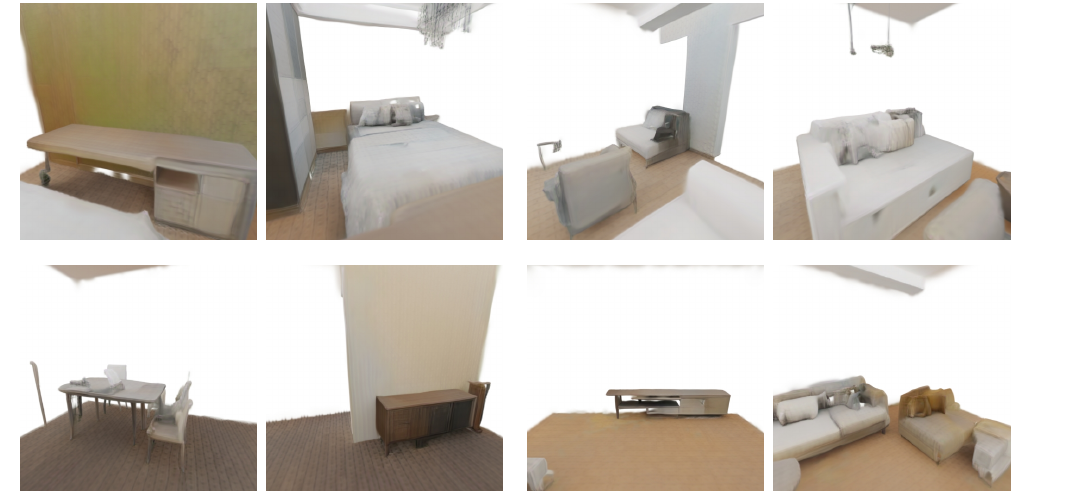} 
  \caption{Qualitative results on unconditional scene chunk generations using L3DG \cite{roessle_l3dg_2024} trained on chunked 3D-FRONT \cite{fu_3d-front_2021} data.}
  \label{fig:l3dg_generations}
\end{figure}

\begin{figure}[htb]
  \centering 
  \includegraphics[width=\textwidth, trim={0cm, 18.9cm, 0cm, 0cm}, clip]{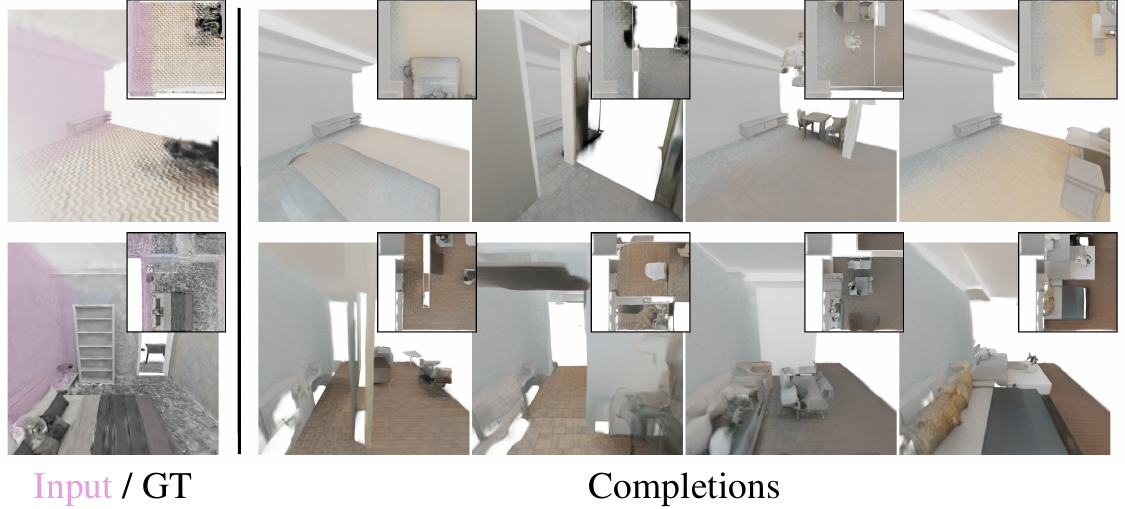} 
  \caption{Qualitative results on scene chunk completion from $25\%$ context using L3DG.}
  \label{fig:l3dg_completions}
\end{figure}

We show qualitative results of L3DG after training on our chunked 3D-FRONT data.
\Cref{fig:l3dg_generations} shows unconditional generation while \cref{fig:l3dg_completions} shows completions.
Compared to GaussianGPT, L3DG respects partial conditions in the completion setting less, as demonstrated by style and geometry changes at the condition boundary.

\newpage
\section{Additional Results} \label{app:additionalresults}
\begin{figure}[bh p]
  \centering
  \includegraphics[width=\textwidth]{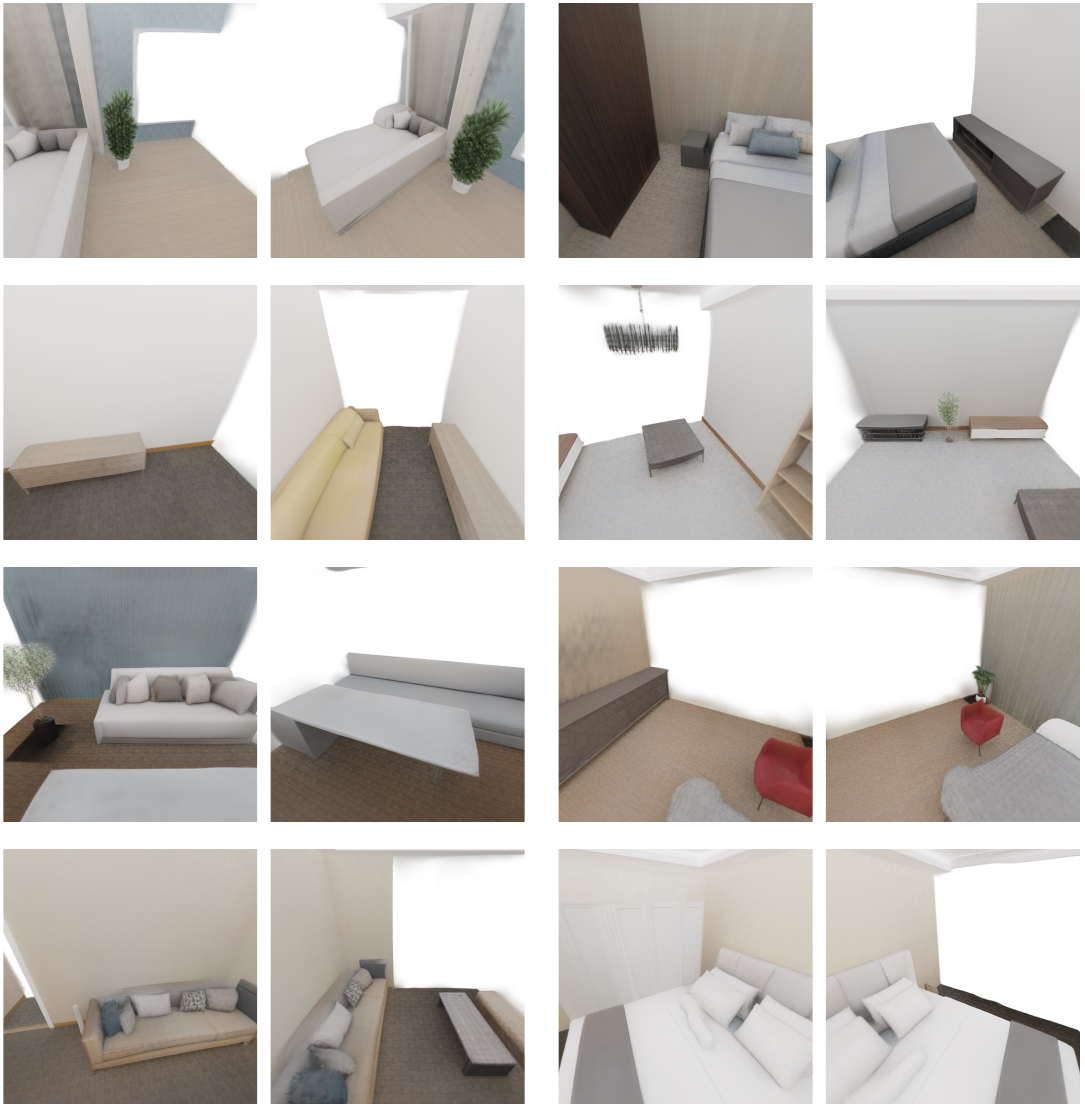} 
  \caption{Additional qualitative results on unconditional scene generation using our model trained only on 3D-FRONT \cite{fu_3d-front_2021}.}
  \label{fig:qual_vfront_app}
\end{figure}

\begin{figure}[p]
  \centering
  \includegraphics[width=\textwidth]{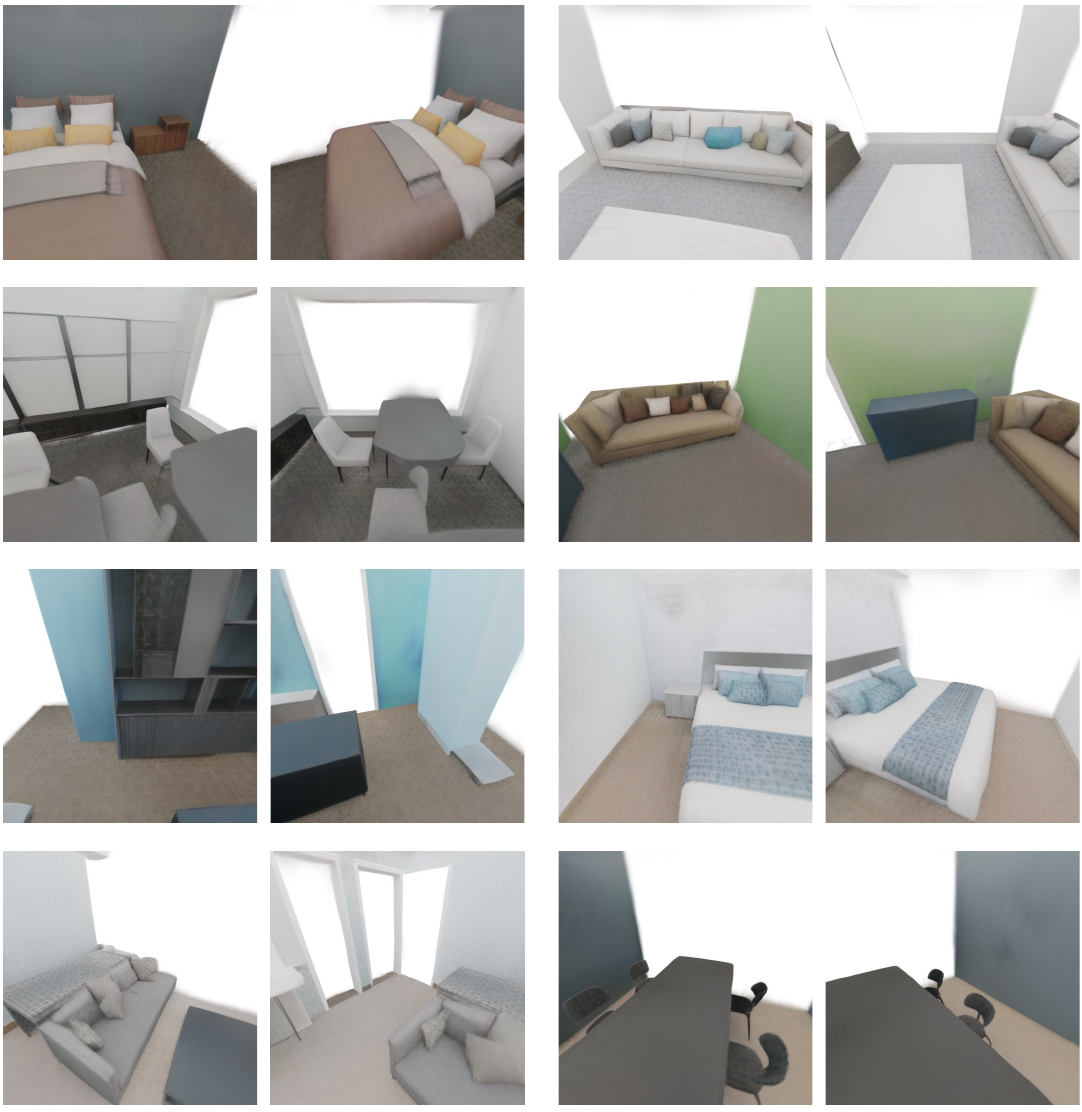} 
  \caption{Qualitative results on unconditional scene generation using our model pre-trained on ASE \cite{avetisyan_scenescript_2024} and 3D-FRONT \cite{fu_3d-front_2021}.}
  \label{fig:qual_vfront_app_ase}
\end{figure}

\begin{figure}[p]
  \centering
  \includegraphics[width=\textwidth, trim={0cm, 4.2cm, 0cm, 0cm}, clip]{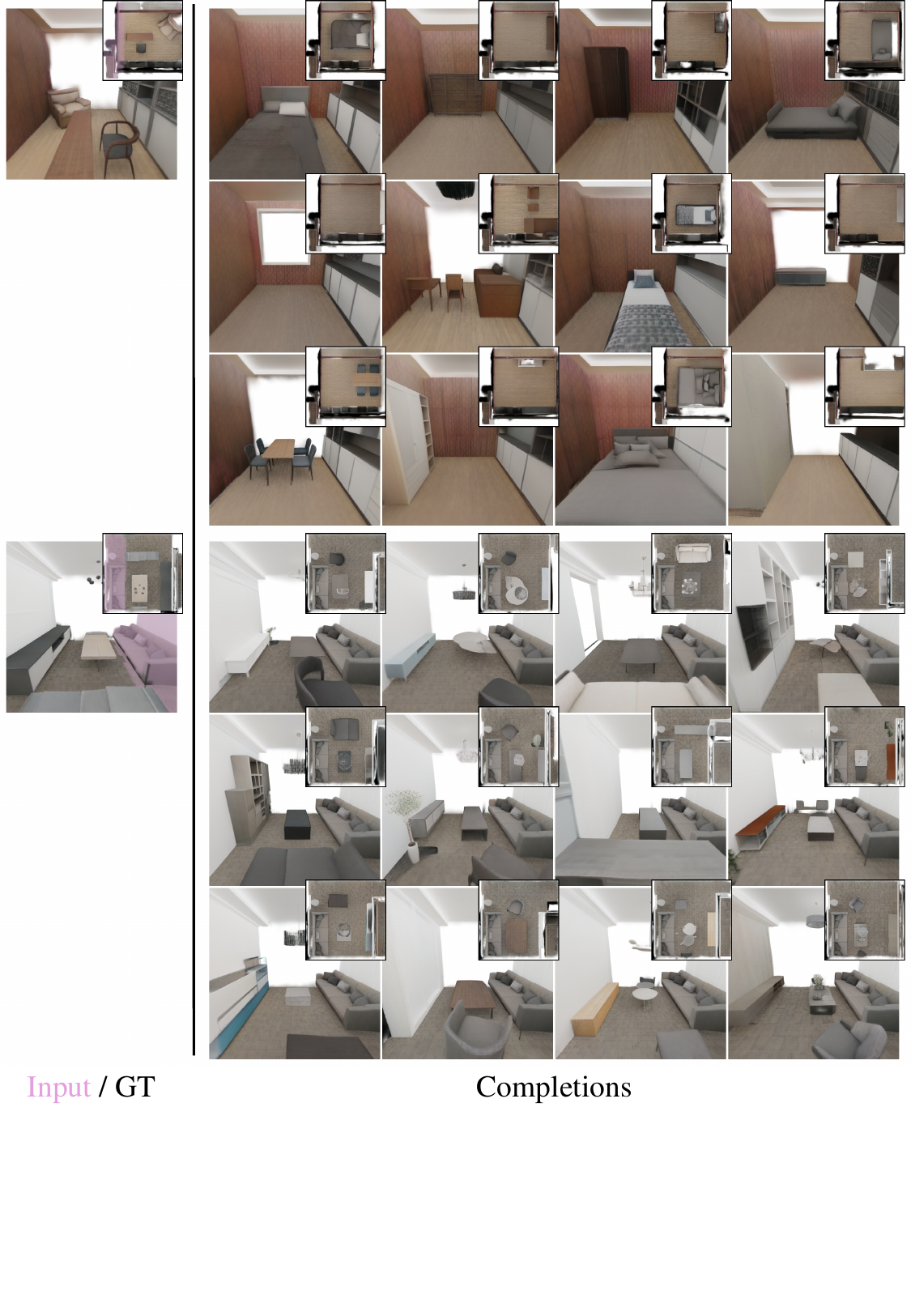} 
  \caption{Additional qualitative results on scene chunk completion.}
  \label{fig:qual_vfront_app_completions}
\end{figure}

We present additional qualitative results of unconditional scene chunk generation using our model trained on 3D-FRONT in \cref{fig:qual_vfront_app}.
\Cref{fig:qual_vfront_app_ase} shows results from our ASE-pretrained model, which improves the diversity of the generated scenes.
Finally, \cref{fig:qual_vfront_app_completions} presents additional scene chunk completions, demonstrating the model's diversity, stability, and quality across a large number of completions per chunk.

\end{document}